\documentclass{article}

\RequirePackage{amsmath}
\RequirePackage[noblocks]{authblk}
\RequirePackage{hyperref} 
\RequirePackage{amssymb}
\RequirePackage{graphicx}
\usepackage[linesnumbered]{algorithm2e}
\let\oldnl\nl% Store \nl in \oldnl

\newcommand{\nonl}{\renewcommand{\nl}{\let\nl\oldnl}}% Remove line
\usepackage{comment}
\usepackage{amssymb}% http://ctan.org/pkg/amssymb
\usepackage{pifont}% http://ctan.org/pkg/pifont
\usepackage{wrapfig}

\newcommand{\cmmnt}[1]{\ignorespaces}  % To comment out some keywords in a text
\graphicspath{ {./figures/av_metadata/}{./figures/synth_samples/}}
\usepackage{framed}
\usepackage{subcaption}
\usepackage{float}

\RequirePackage{fullpage}
\RequirePackage{fancyhdr}

\newcommand{\HRule}{\rule{\linewidth}{0.5mm}}

\makeatletter% since there's an at-sign (@) in the command name
\renewcommand{\@maketitle}{%
  \parindent=0pt% don't indent paragraphs in the title block
  \centering
  {\Large \bfseries\textsc{\@title}}
  \HRule\par%
  \textit{\@author \hfill}
  \par
}

\makeatletter
\def\thickhline{%
  \noalign{\ifnum0=`}\fi\hrule \@height \thickarrayrulewidth \futurelet
   \reserved@a\@xthickhline}
\def\@xthickhline{\ifx\reserved@a\thickhline
               \vskip\doublerulesep
               \vskip-\thickarrayrulewidth
             \fi
      \ifnum0=`{\fi}}
\makeatother

\newlength{\thickarrayrulewidth}
\usepackage{xspace}
\usepackage{xcolor}
\newcommand{\colorcomment}[3]{\xspace{\color{#2}[{#1}]:{#3}}\xspace}
\newcommand{\byung}[1]{\colorcomment{Byung}{red}{#1}}
\newcommand{\ali}[1]{\colorcomment{Ali}{violet}{#1}}

\usepackage[round]{natbib}
\usepackage{multirow}
\makeatother % resets the meaning of the at-sign (@)

\providecommand{\keywords}[1]{\textbf{\textit{Keywords:}} #1}

\usepackage[none]{hyphenat}

\usepackage{array}  

\begin{document}
\sloppy
 
\title{Analysis of Hydrological and Suspended Sediment Events from Mad River Watershed using Multivariate Time Series Clustering}
\author[1]{Ali Javed} 
\author[2]{Scott D. Hamshaw} 
\author[2]{Donna M. Rizzo}
\author[1]{Byung Suk Lee}

\affil[1]{Department of Computer Science, %College of Engineering and Mathematical Sciences, 
University of Vermont, Burlington, VT, USA}
%\affil[2]{GUND Institute for Environment, University of Vermont, Burlington, VT, USA}
%\affil[2]{Vermont EPSCoR, University of Vermont, Burlington, VT, USA}
\affil[2]{Department of Civil \& Environmental Engineering, %College of Engineering and Mathematical Sciences,
University of Vermont, Burlington, VT, USA}

\maketitle % prints the title block

\bigskip
\textbf{Highlights}
\begin{itemize}
    \item This work is the first to apply a computational clustering approach to categorizing and analyzing hydrological storm events defined by multivariate time series.
    \item This work is the first to employ synthetically generated hydrological storm events (engineered to simulate real hydrological storm events) for model validation. 
    \item This work is the first to study the relationship between 3-D time series clustering and 2-D hysteresis loop classification.  
\end{itemize}

\begin{abstract}

Hydrological storm events are a primary driver for transporting water quality constituents such as turbidity, suspended sediments and nutrients. Analyzing the concentration (C) of these water quality constituents in response to increased streamflow discharge (Q), particularly when monitored at high temporal resolution during a hydrological event, helps to characterize the dynamics and flux of such constituents. A conventional approach to storm event analysis is to reduce the C-Q time series to two-dimensional (2-D) hysteresis loops and analyze these 2-D patterns. While effective and informative to some extent, this hysteresis loop approach has limitations because projecting the  C-Q time series onto a 2-D plane obscures detail (e.g., temporal variation) associated with the C-Q relationships. In this paper, we address this issue using a \emph{multivariate time series clustering} approach. Clustering is applied to sequences of river discharge and suspended sediment data (acquired through turbidity-based monitoring) from six watersheds located in the Lake Champlain Basin in the northeastern United States. While clusters of the hydrological storm events using the multivariate time series approach were found to be correlated to 2-D hysteresis loop classifications and watershed locations, the clusters differed from the 2-D hysteresis classifications. Additionally, using available meteorological data associated with storm events, we examine the characteristics of computational clusters of storm events in the study watersheds and identify the features driving the clustering approach.

\end{abstract}

\keywords{Hydrological storm event analysis, streamflow, suspended sediment, clustering, multivariate time series}

\section{Introduction}

Characterizing the rainfall-runoff processes in watersheds is important for understanding the transport of water quality constituents through river systems, the sources of erosion~\citep[e.g.,][]{Sherriff_etal_2016}, and  our ability to evaluate model forecasts~\citep{hess2011}, all of which consequently help with the conservation and management efforts of watersheds~\citep{Bende-Michl2013}.
%Characterizing the rainfall-runoff processes in watersheds is key to understanding the transport of water quality constituents through river systems, which are the spatial sources of erosion~\citep[e.g.,][]{Sherriff_etal_2016}, and to evaluating simulations and forecasts~\citep{hess2011}, both of which are helpful to the conservation and management efforts of watersheds~\citep{Bende-Michl2013}.
Examples of the latter include managing non-point source pollution~\citep[e.g.,][]{Chen_etal_JoH_2017} and monitoring for shifts in watershed function~\citep[e.g.,][]{Burt_etal_HP_2015}. %Characterization of rainfall-runoff processes can be achieved using a machine-learning method --- clustering in our work --- that requires quantifying the degree of similarity (or dissimilarity) between hydrological storm events. \donna{I'm not certain that I understand this last sentence completely...not certain how to re-phrase it. Can this be commented out? It seems a little early in the introduction to be introducing our specific methodology?}

Watershed scientists and environmental managers analyze hydrological data (e.g., response of water quality constituents such as suspended sediment concentration) at the event scale --- in this work, the period of increased storm-runoff response above baseflow as a result of a rainfall event. Constituents are transported primarily during storm events and often show a high degree of variability, for example in the timing of %the 
sediment delivery relative to stream discharge, especially when observed with high frequency monitoring~\citep{Minaudo2017}. Given the variability of both streamflow and water quality constituent responses during hydrological events, it is not surprising that the relationship between such water-quality constituents and discharge are similarly complex and typically cannot be described with simple linear relationships~\citep{Onderka2012}. Despite the added complexity associated with this variation and highly dynamic behavior, the analysis of event concentration-discharge (C-Q) relationships has a long tradition in hydrology, geomorphology, and ecology to infer processes occurring within a watershed~\citep{Aguilera_Melack_2018,Burns_etal_2019,Williams_etal_JoH_2018}.  

A fundamental feature of sediment and solute transport in rivers is that the concentration of such constituents are often not in phase with the associated stream discharge, resulting in hysteresis being present in the C-Q relationship. \cite{WILLIAMS1989} is one of the first to use hysteresis patterns to study hydrological storm events, identifying six classes of hydrological events based on the shape of the hysteresis loops and offering linkages between the hysteresis classes and watershed processes. This hysteresis loop classification continues to be used in present time as a means to grouping storm events~\citep[e.g.,][]{Aguilera_Melack_2018,Rose_etal_HP_2018,Keesstra_etal_2019}. The classification of hysteresis loops is usually done qualitatively using visual patterns~\citep{Hamshaw2018} or quantitatively using a hysteresis index~\citep{Lloyd2016a}. While effective for inferring certain processes, this approach falls short in capturing the full \emph{temporality} of variables, as it ``collapses'' their values as projected on the C-Q plane. The temporality may be seen in the rate of change (e.g., fast, slow), the orientation of change (e.g., clockwise, counter-clockwise), and the shape of change (e.g., linear, convex, concave) in the time series of the C-Q variables. %With the increase in availability of high frequency sensor data, it is now possible to incorporate the temporality of the variables, further refine and add to the existing hysteresis loop classification scheme.%%[Rewritten:]
With high frequency sensor data increasingly available, it is now possible to incorporate the temporality of variables into the analysis, towards further refining and adding to the existing hysteresis loop classification scheme. Additionally, hysteresis loop analysis typically does not consider the degree to which streamflow and suspended sediment return to base conditions at the end of an event - an important characteristic related to antecedent conditions and watershed characteristics.

%there are a limited number of hysteresis loop categories, which often makes the method unsuitable for quantifying the similarity or dissimilarity among storm events within the same hysteresis category.
%\donna{I think we should put a more positive spin on this last sentence. The small number of classifications from Williams may have more to do with the fact that this high resolution data were not available. So I think we should word this more along the lines that, given the advent/availability of high-frequency sensor data, we have the ability to refine and add to this classification scheme.}

%\byung{I moved the discussion on related work to a separate section and broke it into paragraphs. Put it back here if a related work section is not a good idea for JoH.}\ali{It is not conventional to have a separate related work section in JoH and is a part of introduction, also generally true for other hydrology journals. }

%\ali{regroup by single variable studies, and then multiple variable studies. }
%%[Scope #1 of related work]
A few hydrological studies have quantified the similarity between storm events defined by a single variable for categorization or other kinds of modeling (e.g., prediction).
 %
 %%[Details of related work]
 %single series
\cite{hess2011} propose a similarity measure to analyze discharge time series (a.k.a. ``temporal sequence'') that uses feature extraction to leverage attributes of hydrographs such as the rising limb, peak and recession. Such manual feature extraction works well for hydrographs but may not generalize to other water quality time series.
 %
 %single
\cite{EWEN2011178} used a modified version of minimal variance matching (MVM) algorithm~\citep{Latecki2005} to quantify the similarity between storm events. Given a sequence of measurements in a hydrograph (called a ``query sequence''), %minimal variance matching (MVM)
MVM finds a target hydrograph that contains a sub-sequence most similar to the query sequence. This similarity comparison, however, is not symmetric in both directions (i.e., $d(x,y) != d(y,x)$) as MVM can skip some elements of the target sequence~\citep{Latecki2005-2} and, therefore, is not appropriate for use in clustering. 
 %
 %single
\cite{Wendi2019} use cross recurrence plots and recurrence quantification analysis to measure similarity between two hydrographs based on the recurring patterns. Recurrence quantification analysis quantifies the number and duration of recurrences of a dynamic %al
system. Recurrence of subevents is not plausible for the work done at an event scale in this paper. % because the analysis is being done at an event scale.
 %[Scope #2 of related work]
None of these studies, however, was designed for storm events defined by multivariate time series.

In addition, a few other works have applied clustering on storm events defined by multiple variables.
\cite{Bende-Michl2013} used high frequency data to build a database of variables such as precipitation, discharge, runoff coefficient, and maximum discharge, and then performed cluster analysis on these variables to understand nutrient dynamics in the Duck River. 
%Their work includes a limited number of events and uses non-normalized values of discharge and hence results in a focus on categorizing events based on quantities of discharge and phosphorous. 
 %
 %multivariate
\cite{Minaudo2017} studied the relationship between phosphorous and discharge in hysteresis loops by generating high frequency estimates using non-linear modeling~\citep{Jones2011}. They used non-linear regression coefficients to cluster storm events. 
 %
 %multivariate
\cite{MATHER2015} modeled event turbidity as a function of event discharge using a power-law based model. They used cluster analysis on the model parameters to select the number of hysteresis loop categories in developing their classification scheme, thereby avoiding the use of predetermined classes. %discharge and turbudity. multivariate
None of these works, however, categorizes storm events by capturing the full temporality of variables as defined by the rate of change, the orientation of change and the shape of change in the time series of C-Q variables.

%One fundamental issue in this project is that the current hysteresis analysis does not fully utilize the \emph{temporality} of the event and, therefore, a 3-D plot that includes the time axis, namely a spatiotemporal trajectory, is needed. The temporality may be seen in the rate of change (e.g., fast, slow), the orientation of change (e.g., clockwise, counterclockwise), and the shape of change (e.g., linear, convex, concave).

In this paper, we present a method to cluster multivariate water quality time series at the event scale. As an example, we use multivariate time series clustering on two variables: concentration (C) and discharge (Q) by modeling them as trajectories %of captured signals (in short, ``signal trajectories'')
in a 3-D space defined by concentration, discharge, and time, i.e., C-Q-T plane. We use high-resolution riverine suspended-sediment concentration (SSC) time series -- hereafter referred to simply as concentration -- collected from six watershed sites in Vermont %\donna{insert \# rather than saying "multiple"?}
for up to three years and show proof-of-concept of applying the computational clustering methods to categorize hydrological storm events. %The method presented in this work is automated and unsupervised. 
The efficacy of the approach is demonstrated qualitatively using multi-dimensional event visuals and quantitatively using metrics that summarize event characteristics. 
%We also discuss the implications of our work for other Vermont watersheds.\ali{We have so far not discussed the implications so if we do not do that, remove this line.}

\section{Study Area and Data}\label{sec:studyArea}

\begin{wrapfigure}[17]{r}{0.5\linewidth}
\centering
\vspace*{-1em}
\includegraphics[width=0.5\textwidth]{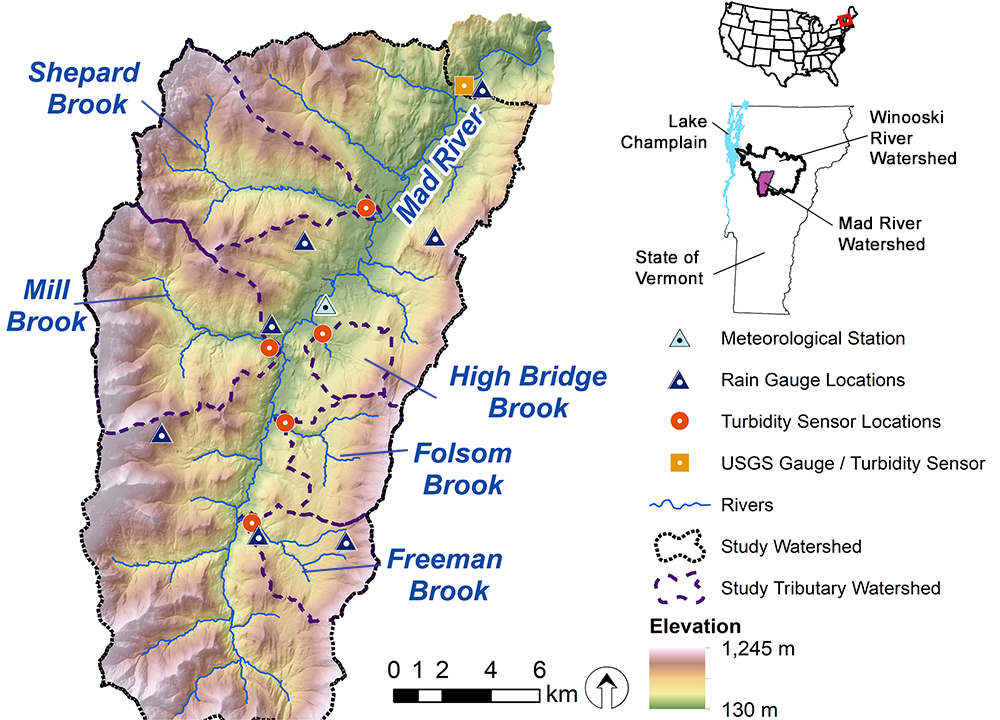}
\caption{Study area locations within the Lake Champlain Basin of Vermont~\citep{Hamshaw2018}.}
\label{fig:Lake_Champlain}
\end{wrapfigure}

Our study area is located in the Mad River watershed in the Lake Champlain Basin and central Green Mountains of Vermont (see Figure~\ref{fig:Lake_Champlain}). 
This area was selected primarily due to the availability of continuous streamflow and suspended sediment monitoring data~\citep{Hamshaw2018} and ongoing geomorphic and sediment dynamics studies at the University of Vermont~\citep{Stryker2017,Wemple2017}. Data of more than 600 storm events were collected in this watershed (and its five sub-watersehds) from October $19^{th}$, 2012 to August $21^{th}, 2016$ (see Table~\ref{tbl:watershed_eventcount}). ~\cite{Hamshaw2018} previously used this dataset to automate the C-Q hysteresis loop classification and further refine the hysteresis classification of \cite{WILLIAMS1989}. Sensors were installed to gather discharge and turbidity data at 15\,minute intervals at each of the five tributaries and the main stem shown in Figure~\ref{fig:Lake_Champlain}. Suspended sediment concentration (SSC) was estimated from turbidity using regression relationships. Storm event identification from the continuous stream of data was done in a semi-automated fashion. Mad River watershed storm events also have associated meteorological data available as summarized in the 24 storm event metrics (see Table~\ref{tbl:performance_metrics}). Details on data collection and event pre-processing in the Mad River watershed can be found in \cite{Hamshaw2018}.

\begin{table}[htp!]
\caption{Number of storm events from each study watershed.
}\label{tbl:watershed_eventcount}
\centering
\newcolumntype{M}[1]{>{\centering\arraybackslash}m{#1}}
\begin{tabular}{|c|M{2.6cm}|M{2.6cm}|M{2.6cm}|}
\hline

{\bf Site}& {\bf Number of events monitored}& \bf{Monitoring start date} & \bf{Monitoring end date} \\
\hline
Mad River (main stem) & 148 & Oct $29^{th}, 2012$ & Aug $21^{th}, 2016$\\
Shepard Brook & 106 &  Jul $18^{th}, 2013$ & Dec $23^{rd}, 2015$\\
High Bridge Brook & 41 & Jun $6^{th}, 2013$ & Nov $17^{th}, 2013$\\
Mill Brook & 158 & Oct $19^{th}, 2012$ & Dec $23^{rd}, 2015$ \\
Folsom Brook & 96 & Jul $17^{th}, 2013$ & Sept $13^{th}, 2015$ \\
Freeman Brook & 54 & Jun $2^{nd}, 2013$ & Nov $17^{th}, 2013$ \\
\hline
{\bf Total} & \bf{603} & \bf{Oct \boldmath$19^{th}, 2012$} & \bf{Aug \boldmath$21^{th}, 2016$} \\
\hline
\end{tabular}

\end{table}

\begin{table}[!htp]

\caption{Description of the 24 storm event metrics used in this work.}\label{tbl:performance_metrics}
\centering
\begin{tabular}{|c|c|}
\hline
{\bf Metric}&{\bf Description}\\
\hline
\multicolumn{2}{|c|}{\bf Hydrograph/ Sedigraph characteristics}\\
\hline
${T}_{Q}$& Time to peak discharge (hr)\\
${T}_{SSC}$& Time to peak TSS (hr)\\
${T}_{QSSC}$& Time between peak SSC and peak flow (hr)\\
$Q_{Recess}$ & Difference in discharge value at the beginning and end of event\\
$SSC_{Recess}$ & Difference in concentration value at the beginning and end of event\\
HI& Hysteresis Index\\
\hline
\multicolumn{2}{|c|}{\bf Antecedent conditions}\\
\hline
$T_{LASTP}$&Time since last event (hr)\\
A3P& 3-Day antecedent precipitation (mm)\\
A14P& 14-Day antecedent precipitation (mm)\\
${SM}_{SHALLOW}$& Antecedent soil moisture at 10 cm depth (\%)\\
$SM_{DEEP}$& Antecedent soil moisture at 50 cm depth (\%)\\
$BF_{NORM}$& Drainage area normalized pre-storm baseflow ($m^3/s/km^2)$\\
\hline
\multicolumn{2}{|c|}{\bf Rainfall characteristics}\\
\hline
P& Total event precipitation (mm)\\
$P_{max}$ & Maximum rainfall intensity (mm) \\
$D_{P}$& Duration of precipitation (hr)\\
${T}_{PSSC}$& Time between peak SSC and rainfall center of mass (hr) \\
\hline
\multicolumn{2}{|c|}{\bf Streamflow and sediment characteristics}\\
\hline
BL & Basin Lag\\
${Q}_{NORM}$& Drainage area normalized stormflow ($m^3 / s / km^2$)\\
Log(${Q}_{NORM}$)& Log-normal stormflow quantile (\%)\\
${D}_{Q}$& Duration of stormflow (hr)\\
FI& Flood intensity\\
SSC& Peak SSC (mg/L)\\
${SSL}_{NORM}$& Drainage area normalized total sediment ($kg/m^2$)\\
${FLUX}_{NORM}$& Drainage area and flow normalized sediment flux ($kg/m^3/km^2$)\\
\hline
\end{tabular}
\end{table}

The elevation of the Mad River watershed ranges from 132\,m to 1,245\,m above sea level, and is predominantly forested except for the valley bottom, which features agriculture, village centers, and other developed lands (see Table~\ref{tbl:watershedCharacteristics}). The watershed has a mean annual precipitation ranging from approximately 1,100\, mm along the valley floor to 1,500\,mm along the upper watershed slopes~\citep{PRISM2015}. Soils range from fine sandy loams derived from glacial till deposits in the uplands to silty loams derived from glacial lacustrine deposits in the lowlands. Erosional watershed processess include bank erosion, agricultural runoff, unpaved road erosion, urban storm water, and hillslope erosion. Similar to many watersheds in Vermont, reducing excessive erosion and sediment transport in the Mad River is a focus of the management efforts such as implementation of stormwater management practices, streambank stabilization, and river conservation. % among others. 

\begin{table}[!htb]
\caption{Key characteristics of the study watersheds.}
\label{tbl:watershedCharacteristics}
\newcolumntype{M}[1]{>{\centering\arraybackslash}m{#1}}
\newcolumntype{C}[1]{%
 >{\vbox to 2ex\bgroup\vfill\centering\arraybackslash}%
 m{#1}%
 <{\egroup}}  
%\newcolumntype{M}[1]{>{\centering\arraybackslash}m{#1}}
\begin{small}
\resizebox{\columnwidth}{!}{%
%\resizebox{\textwidth}{!}{%
\begin{subtable}{1\textwidth}

\centering
\begin{tabular}{|C{2.39cm}|p{1.3cm}|p{1.0cm}|p{0.9cm}|p{1.1cm}|p{1.3cm}|p{0.9cm}|}

%\begin{tabular}{|c|c|c|c|c|c|c|c|c|c|}
\hline
{\bf Characteristic} & {\bf Shepard Brook} & {\bf High Bridge Brook} & {\bf Mill Brook} & {\bf Folsom Brook} & {\bf Freeman Brook} & {\bf Mad River} \\
\hline
Area ($km^2$) & 44.6 & 8.6 & 49.2 & 18.2 & 17.0 & 344.0 \tabularnewline
Minimum elevation (m) & 195 & 225 & 216 & 229 & 266 & 140 \tabularnewline 
Maximum elevation (m) & 1117 & 796 & 1114 & 886 & 860 & 1245\tabularnewline
Elevation range (m) & 923 & 571 & 898 & 657 & 594 & 1105 \tabularnewline
Stream order & 4th & 3rd & 4th & 4th & 4th & 5th \tabularnewline
Drainage density ($km/km^2$) & 2.38 & 2.45 & 2.16 & 1.77 & 1.95 & 0.97 \tabularnewline
\% Forested land & 92.2 & 66.7 & 89.2 & 77.6 & 76.2 & 85.5\tabularnewline
\% Developed land & 1.0 & 16.6 & 1.5 & 12.7 & 8.3 & 4.7\tabularnewline
\% Agricultural land & 5.6 & 15.5 & 7.0 & 8.8 & 14.6 & 8.0 \tabularnewline
\% Other land & 1.1 & 2.1 & 0.8 & 0.7 & 1.7 & 1.1\tabularnewline

\hline
\end{tabular}
\end{subtable}
}
\end{small}
\end{table}

\section{Methods}\label{sec:methods}

\subsection{Event time series processing}\label{sec:preprocess}

%In this work, sensor data collected for individual storm events are modeled as a ``signal trajectoriesy' and are mathematically represented as multivariate (bi-variate, in the case of C-Q)\donna{Should we keep this generic early on?} time series. A time series is a sequence of data points ordered by time and typically observed at regular intervals (e.g., 15-minutes) in environmental applications. \donna{Is "in environmental applications" needed?} A multivariate time series represents multiple (univariate) time series simultaneously by combining their data points into one vector point. For example, a time series of streamflow and a time series of SSC are combined into a time series of $\langle\text{streamflow}, \text{SSC}\rangle$, a (mathematical) vector of the two.

%In this work, sensor data collected for individual storm events are modeled as ``signal trajectories' and are mathematically represented as multivariate (bi-variate, in the case of C-Q)\donna{Should we keep this generic early on?} time series. A time series is a sequence of data points ordered by time and typically observed at regular intervals (e.g., 15-minutes). A multivariate time series combines one or more univariate time series into a matrix (where the number of columns represent the number of univariate time series data under consideration and the number of rows equals the duration (time, t) of the event.\donna{$\leftarrow$Did I change the meaning?} For example, a time series of  streamflow and SSC data would be combined into a matrix with dimensions (t x 2).

\begin{wrapfigure}[14]{r}{0.4\linewidth}
\centering
%\vspace*{-1em}
\includegraphics[width=0.3\textwidth]{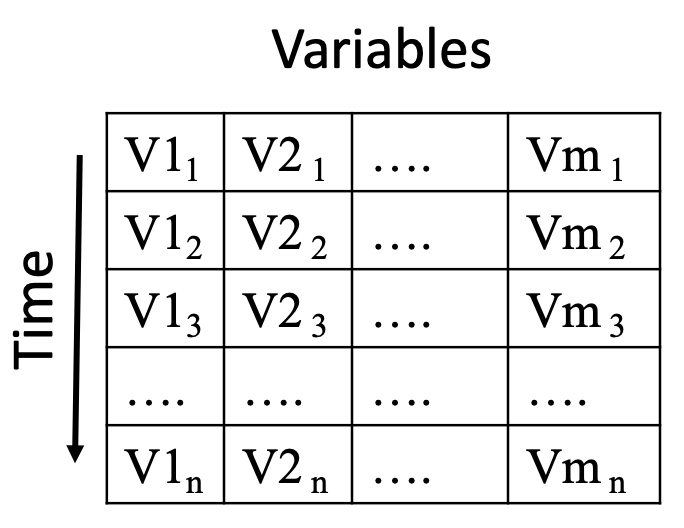}\vspace*{-1em}\\
\caption{A matrix representation of multivariate time series ($m$ variables, $n$ time steps); a column for each variable and a row for variable value at each time step.}
\label{fig:timeseries_example}
\end{wrapfigure} 

A time series is a sequence of variable values ordered by time and typically observed at a regular interval. In this work, sensor data collected for individual storm events are modeled as %``signal trajectories'' 
trajectories and are mathematically represented as multivariate time series. A \emph{multivariate} time series is a times series of two or more variables combined. For example, two (univariate) time series, $T1 = \langle V1_1, V1_2, V1_3, ..., V1_n\rangle$ and $T2 = \langle V2_1, V2_2, V2_3 ..., V2_n\rangle$, when combined, make a bivariate time series $\mathbf{T} = \langle (V1_1, V2_1), (V1_2, V2_2), ..., (V1_n, V2_n)\rangle$. %where the two variables are $V1$ and $V2$ 
(See Figure~\ref{fig:timeseries_example} for a matrix representation of a multivariate time series with $m$ variables and $n$ time steps.). 
%\byung{How about adding a figure illustrating the mapping of a trajectory time series of length $n$ being mapped to a point in an $n$-dimensional space and explaining that multivariate time series clustering is done on those points?} \ali{figure referred to later too.}

\begin{figure}%[!ht]
\centering
\begin{subfigure}[t]{0.22\textwidth}
\centering
\caption{}\label{fig:no_processed_sediment}
\includegraphics[width=1\textwidth]{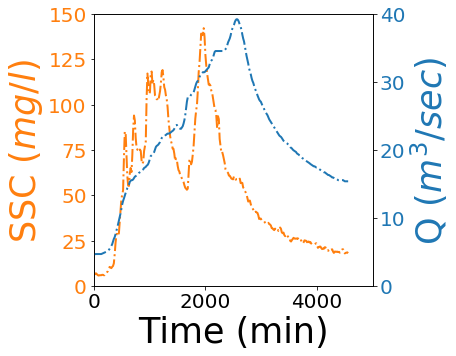}
\end{subfigure}\qquad
\begin{subfigure}[t]{0.20\textwidth}
\centering
\caption{}\label{fig:no_processed_stream}
\includegraphics[width=1\textwidth]{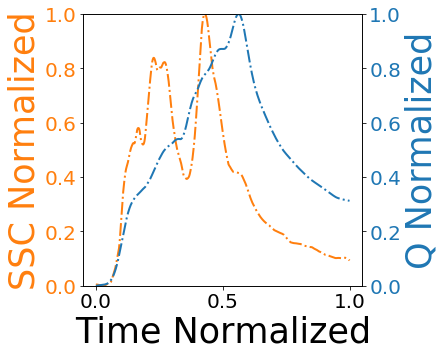}
\end{subfigure}\qquad
\begin{subfigure}[t]{0.17\textwidth}
\centering
\caption{}\label{fig:normalized_cq}
\includegraphics[width=1\textwidth]{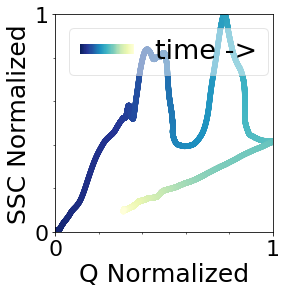}
\end{subfigure}\qquad
\begin{subfigure}[t]{0.18\textwidth}
\centering
\caption{}\label{fig:normalized_cqt}
\includegraphics[width=1\textwidth]{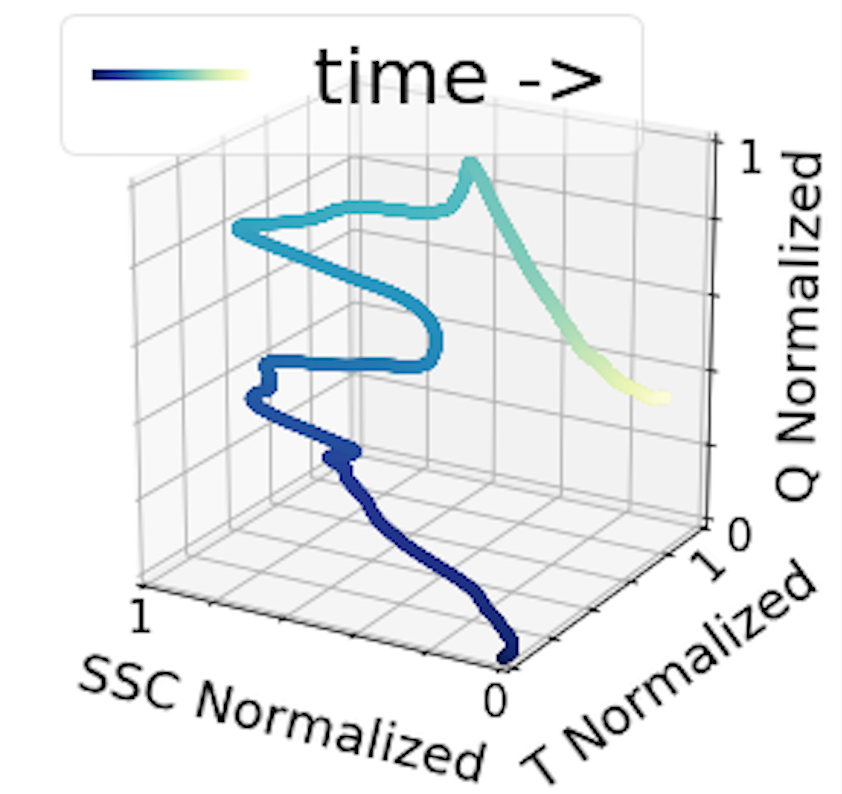}
\end{subfigure}
\caption{Pre-processing %and normalization 
of storm events: (a) raw time series, (b) %Smoothed and normalized 
pre-processed time series, (c) C-Q plot, and (d) C-Q-T plot.}\label{fig:preprocessing}
\end{figure}

The environmental time series data in our work are collected in-situ by multiple sensors. These data typically contain noise and gaps, and therefore pre-processing (i.e., filtering and re-sampling) are often necessary.
%they are removed (or reduced) through %filtering 
%smoothing and re-sampling.  
%
In addition, given that the data are delineated into %by
hydrological events and our %goal is to compare event dynamics in terms of 
interest is in comparing the relationship between discharge and concentration, %\donna{in space and time?},
we normalized both of them %discharge and concentration
within each event to facilitate the study of C-Q and C-Q-T plots (see Figure~\ref{fig:preprocessing}).
%
%\begin{wrapfigure}{l}{0.5\linewidth}[ht]
%\end{wrapfigure}
 %
Specifics of these pre-processing steps performed are as follows:
\begin{description}
    \item[Smoothing.] Discharge and concentration time series were smoothed to remove noise using the Savitsky-Golay Filter~\citep{Savitzky-Golay}. We selected a third-order, 21-step filter for the main stem of Mad River and a fourth-order, 13-step filter for the remaining sub-watersheds. %oth the choice of filter order and step size preserved the peaks, overall shape and were based on visual inspection of event time series, consistent with previous work by~\cite{Hamshaw2018}.
    Both choices of the filter order \& step size were based on visual inspection of the resulting event time series to preserve the peaks and overall shapes, in the same manner as was done in the previous work by~\cite{Hamshaw2018}. 
    
    \item[Re-sampling.] Discharge and concentration time series were re-sampled to a uniform %event 
    length of 50 %data points
    samples using univariate spline fitting~\citep{Univariate-Spline}. 
    %In other words, the temporal dimension of each time series was normalized to 50 points (empirically selected) to 
    The length 50 was selected empirically as the minimum possible length that preserves the shape and characteristics of the event time-series. The re-sampling ensures that clustering is affected not by the length of the event but by the shape of the %resulting signal 
    trajectory.
    \item[Normalization.] %Data for each event was normalized to keep the magnitudes of discharge and concentration between 0 and 1. 
    Discharge and concentration time series were scaled to the range of 0 to 1 in their magnitude. 
    %Magnitude of samples in discharge and concentration time series were scaled to the range of 0 to 1. 
    This normalization ensures that the clustering is affected not by the magnitude of the individual times series but by the shape of the trajectory. (Normalization of magnitude is commonly used for a meaningful comparison of time series~\citep{Rakthanmanon2012}.)
\end{description}

\subsection{Concentration-discharge (C-Q) hysteresis classification}\label{sec:hysteresis_loops}

%\begin{wrapfigure}{r}{0.6\linewidth}
\begin{figure}[!hb]
\centering
\includegraphics[width=1.0\textwidth]{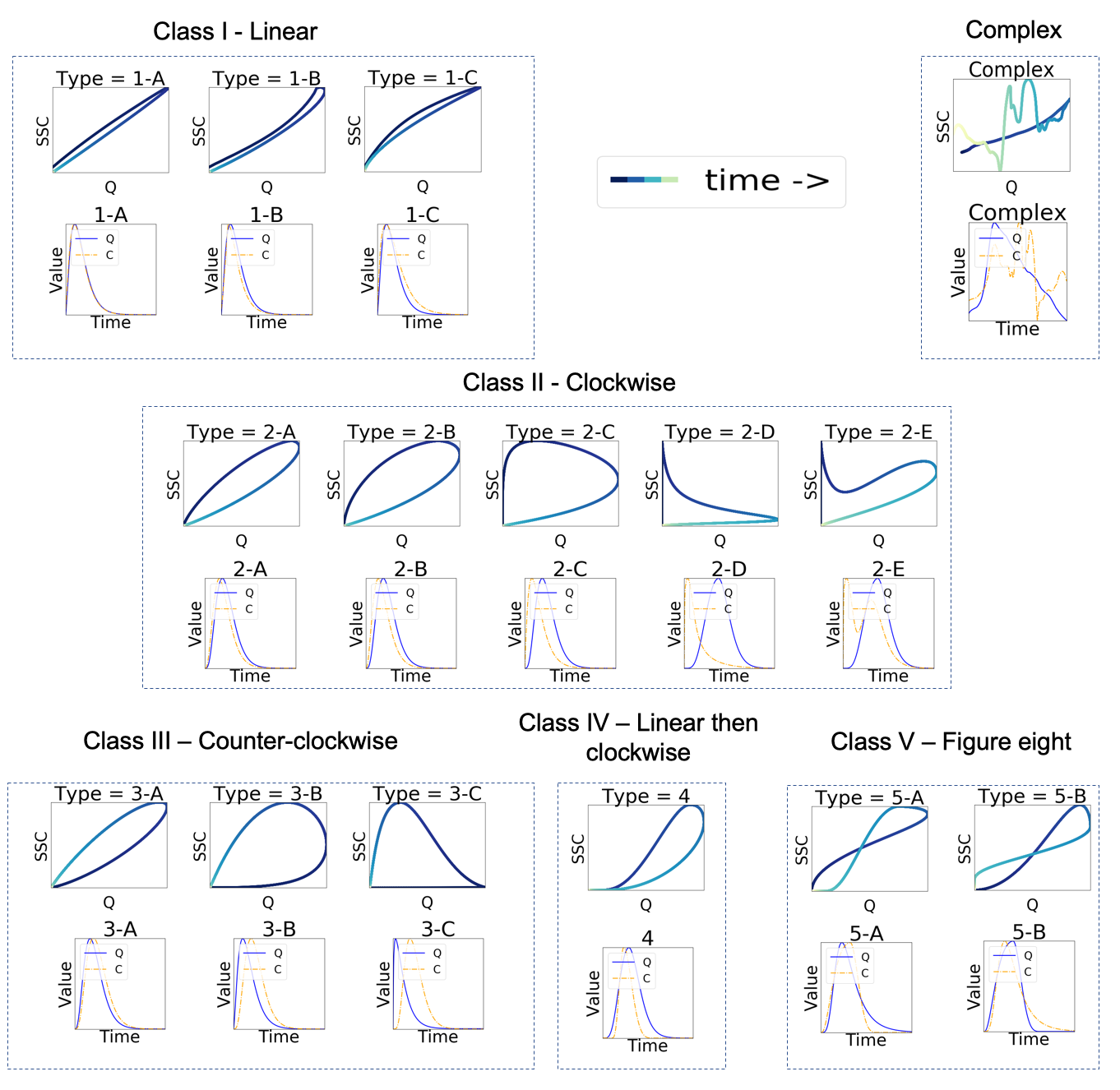}\\
\caption{
Classes of discharge-SSC hysteresis loops from events observed in the Mad River watershed. Solid line indicates hydrograph and dashed line indicates sedigraph.}
\label{fig:hysteresisLoop}
\end{figure}
%\end{wrapfigure}
%
Each event in our dataset was %classified 
categorized visually (by a human) in %to 
two %different 
classification schemes (see Figure~\ref{fig:hysteresisLoop}): %the
six classes of~\cite{WILLIAMS1989} and expanded 15  sub-classes of~\cite{Hamshaw2018}. We refer to categorization by Williams as ``Classes'' and the subcategories by Hamshaw as ``Types'' in this work. %\ali{Since the naming convention is constantly disputed, added this sentence to highlight how the naming is used in this draft} \scott{Don't change for preprint to CoRR, but I think when we go to a journal, we should change to calling Traditional hysteresis classes and Expanded or similar}. 
Class I and its subcategories represents 
%variations on the 
C-Q relationships that show no hysteretic behavior. %Class II and III represent clockwise and counter-clockwise hysteretic behavior, respectively; and the subcategories are differentiated by timing of the peak discharge and peak SSC influencing the shape of the hysteresis. 
Class II represents those with clockwise hysteretic behavior and Class III with counter-clockwise, and their subcategories are differentiated by timing of the peak discharge and peak SSC influencing the shape of the hysteresis. 
A C-Q plot with a linear relationship followed by a clockwise loop is indicative of Class IV behavior; these patterns could reasonably be considered a special case of Class II (clockwise) hysteresis patterns, since they are linear first and then clockwise. The figure-eight shaped loops are represented as Class V, with subcategories discriminated %between
by the loop orientation. Events that do not fall into any of the %categories
classes above were placed into the hysteresis %pattern 
class labeled ``complex''.

\subsection{Multivariate time series clustering}\label{sec:multi-clustering}

%\ali{Move the partitioning comment later in the draft because method has not been explained yet.} 
We clustered our multivariate time series data at the event scale into groups that correspond %\donna{I'm still not certain I follow the remainder of this sentence; let's talk further about this after we see the temporal anaysis for one watershed. We clustering over time and space?} 
to various hydrograph and water quality (e.g., sedigraph) characteristics. To this end, significant effort was made to choose the clustering method. \cite{Paparrizos2016,Paparrizos2017} conducted extensive benchmark tests on different clustering algorithms using multiple datasets from University of California at Riverside (UCR) time series repository~\citep{UCRArchive2018}
%
%\cite{Paparrizos2016} conducted detailed benchmark tests on six clustering algorithms and three different distance measures using 48 of the benchmark datasets in the University of California at Riverside (UCR) time series repository~\citep{UCRArchive2018} 
and found K-medoids with dynamic time warping (discussed in Section~\ref{sec:dtw}) to be the most accurate. Leveraging their work, we conducted additional benchmark tests on four different algorithms --- TADPole~\citep{Nurjahan2016}, Kshape~\citep{Paparrizos2016}, K-medoids (Dynamic time warping), K-medoids (Euclidean) ---  using all 128 datasets currently present in the UCR time series repository~\citep{UCRArchive2018} and 
found K-medoids with dynamic time warping to be most accurate (highest average Rand Index). All event times series data were pre-processed as outlined in Section~\ref{sec:preprocess}.

%\donna{Did I change the meaning? Note: That later you say the four algorithms are described in this section; but I'm not so certain that's the case. Are we really comparing 4 different clustering algorithms? Again, let's talk this over.}\byung{Ali and I are looking at this comment. The draft mentions ``four selected algorithms'' later in Section 4.1. A decision was made early on that the algorithms are not to be discussed, as it may not be of interest to the JoH reviewers. If we need to, however, we can do that.} 

\subsubsection{K-medoids clustering algorithm}\label{sec:kmed}

K-medoids is a variant of the popular K-means~\citep{Wu2007}, where the cluster centroids are actual data points (called ``medoids'') as opposed to coordinates as in K-means. These data points are in an $n$-dimensional space mapped from a multivariate time series of length $n$, where at each time step is a vector of the multiple variables (e.g., $V1, V2, ..., Vm$ in Figure~\ref{fig:timeseries_example}). Like K-means, K-medoids is an iterative algorithm (see Algorithm~\ref{alg:k-medoids}) where the initial centroids are randomly selected. The algorithm comprises two phases: the phase 1 assigns data points to clusters (Line~\ref{lin:associate} of Algorithm~\ref{alg:k-medoids}) and the phase 2 calculates new centroids for each cluster (Line~\ref{lin:center_calc} of Algorithm~\ref{alg:k-medoids}). In the first phase, the distance between all data points and each of the centroid is calculated, and each data point is assigned to the closest centroid. In the second phase, a new centroid is selected from each cluster by finding the data point that minimizes the sum of distances from it to all other data points in the cluster (called the ``cost of configuration''). These two phases are repeated for a fixed number of times or until there is no change in the centroid selection. Algorithm~\ref{alg:k-medoids} was implemented in Python (version 3.6.1); the source code can be found at \cite{multi_kmed}.

\begin{algorithm}[ht]
\caption{K-medoids algorithm for storm event clustering.}\label{alg:k-medoids}
\nonl \underline{Algorithm K-medoids}\\
\SetAlgoLined
\nonl Input: storm events (i.e., their multivariate time series  representations); number $k$ of clusters to be generated\\
\nonl Output: $k$ clusters generated from the events
\smallskip\\
\nonl {Procedure}% ($Tweets$) 
\smallskip\\
// Initialize random seeds.\\
Randomly select $k$ events as medoids from the input events. \label{lin:random_seeds}\\
\While{termination criterion is not met}{
    // Termination condition can be convergence of medoids or maximum allowed iterations.\\
    %\smallskip\\
    Phase 1: Assign each event to its closest medoid.\label{lin:associate}\\
    Phase 2: From each cluster consisting of the medoid and events assigned to it, select an event that gives the smallest sum of distances to all the other events in the cluster and make the selected event a new medoid. \label{lin:center_calc}
}

Return each cluster consisting of a medoid and all other events assigned to it.

\end{algorithm}

\subsubsection{Dynamic time warping}\label{sec:dtw}

In order to cluster storm events represented by concentration and discharge time series, we used a variant of dynamic time warping (DTW) to calculate the ``distance'' %metric \ali{removed the word metric, because DTW is not a "distance metric" since it violates triangular inequality, we do not need to talk about that here, but shouldnt use the word metric. }
between two multivariate times series representing different 
storm events. Originally introduced for speech recognition~\citep{Sakoe1978}, DTW is now arguably the most popular distance measure for time series data and is particularly appealing for data generated in the hydrological environment because of (i) the difficulty in defining the beginning and end of a hydrological event (i.e., the ambiguity inherent in event delineation), and (ii) the presence of noise in the sensor data (e.g., fouling). 

%%Euclidean distance (Equation~\ref{eq:euc}\byung{This equation may be too trivial. Omit it?}\ali{While it is trivial I am building up on this later hence for clarity maybe it should be here}) 
%\begin{equation}
%d(\mathbf{p},\mathbf{q}) = \sqrt{\sum_i^n (q_i %-p_i)^2}\label{eq:euc}
%\end{equation}
%\noindent where $d(\mathbf{p}, \mathbf{q})$ is the distance between two trajectories $\mathbf{p} (= p_1,p_2,...,p_n)$ and $\mathbf{q} (= q_1, q_2, ..., q_n)$.

\begin{figure}[!ht]
\centering
\begin{subfigure}[b]{0.40\textwidth}
\centering
\caption{}
\includegraphics[width=1\textwidth]{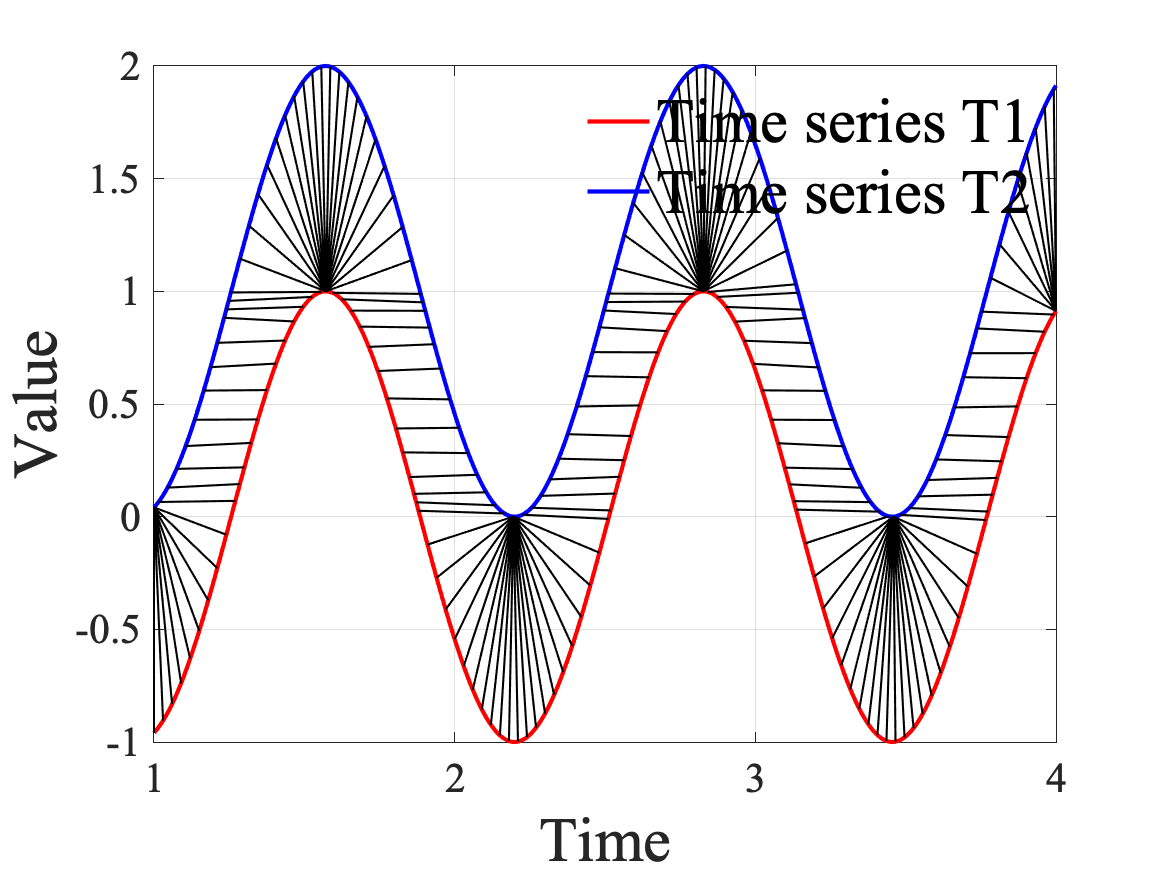}\quad
\end{subfigure}
\begin{subfigure}[b]{0.40\textwidth}
\centering
\caption{}
\includegraphics[width=1\textwidth]{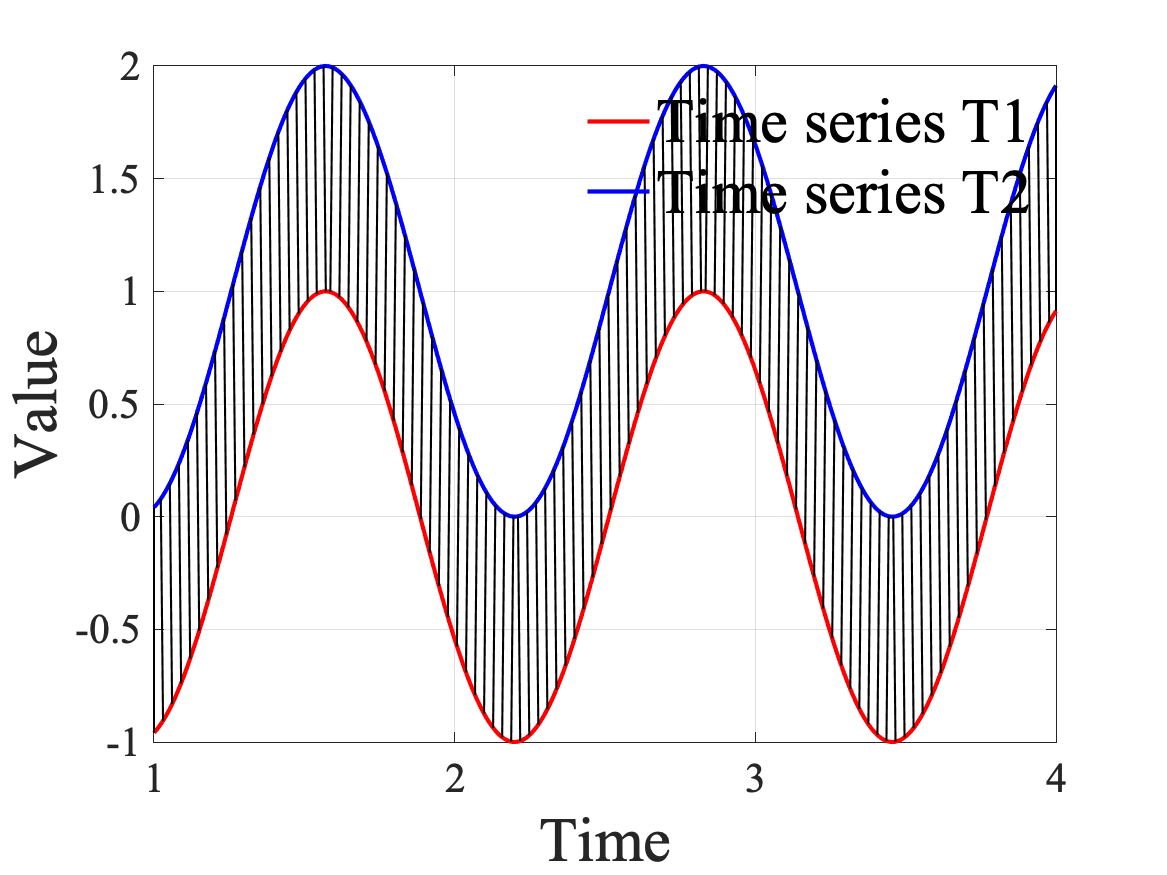}%\quad
\end{subfigure}
\caption{Illustration of the alignment between two times-series for calculating distance in (a) dynamic time warping (one-to-many) and (b) Euclidean (one-to-one).}
\label{fig:DTW_EUC}
\end{figure}

Figure~\ref{fig:DTW_EUC} illustrates how the distance is calculated between two \cmmnt{univariate} time series (blue and red) using DTW compared to the more common Euclidean distance. While the Euclidean distance metric uses a one-to-one alignment; DTW  employs a one-to-many alignment.  This one-to-many alignment allows DTW to warp the time dimension so as to minimize the distance between the two time series. DTW can optimize alignment, both global alignment (by shifting the entire time series left or right) and local alignment (by stretching or squeezing sections of time series). This warping is often constrained to a limited neighborhood defined by a window. Experiments conducted by \cite{Paparrizos2016} showed the best accuracy (as measured by Rand Index) was obtained by constraining DTW to a limited window. Our intuition is that such a constraint results in better accuracy on average since too much flexibility may result in falsely high similarity values. Moreover, constraining the window size to 10\% of the data is usually considered more than adequate for real-world applications~\citep{chotirat2004}. We also use this 10\% window constraint in our calculation of DTW, because it allows flexibility in approximating the beginning and end of a hydrological event. DTW-D was implemented in Python (version 3.6.1); the source code can be found at \cite{dtw_d_code}.% and is likely to reflect the timing and shape of events better in the clustering results. % aligns with our desire to bias the clustering (i.e., matching of the event's hydrograph/sedigraph times series) based on the overall timing and shape of the event.

%desired to maintain resiliency to misidentified start/end points and some distortions in an event's hydrograph/sedigraph so that matching can be done based on the overall timing and shape of events. As shown in 

\begin{figure}%[!ht]
\centering
\begin{subfigure}[b]{0.30\textwidth}
    \centering
    \caption{}\label{fig:matrix}
    \includegraphics[width=1\textwidth]{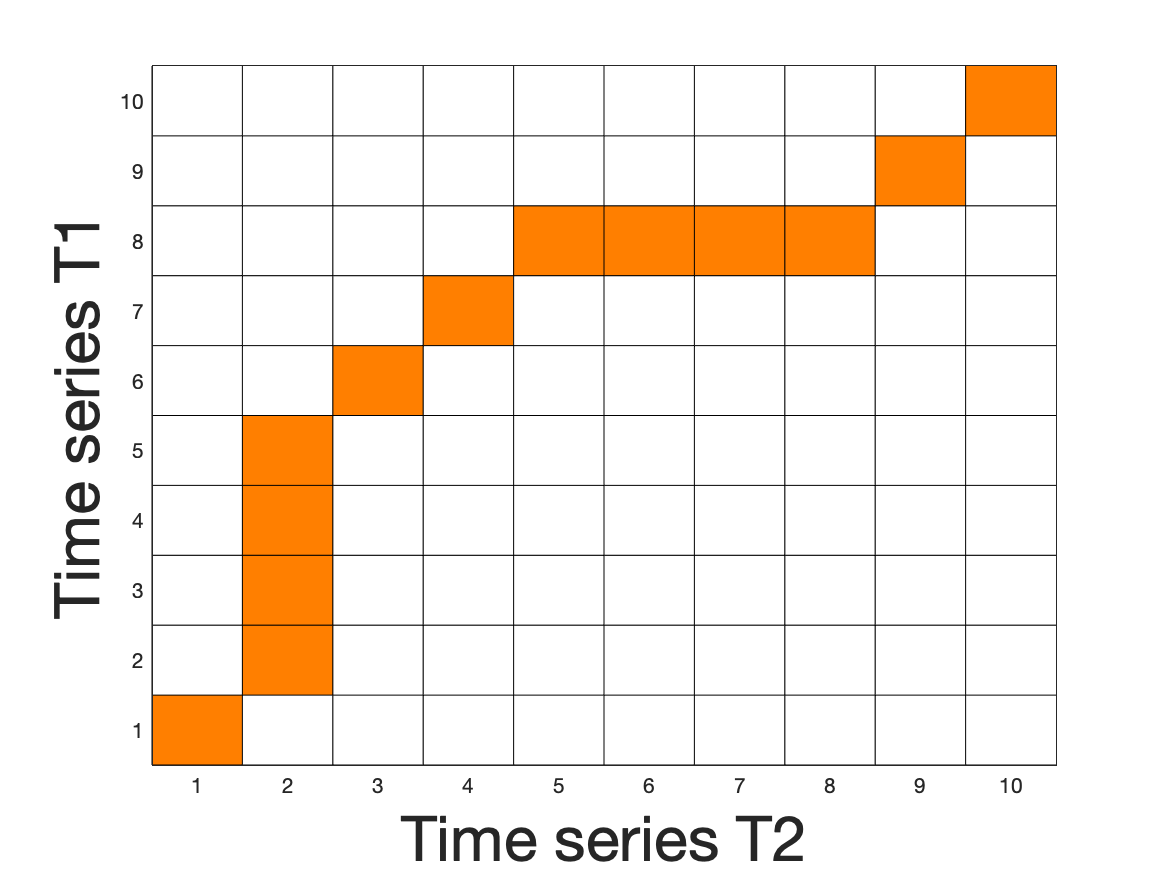}
\end{subfigure}
\begin{subfigure}[b]{0.30\textwidth}
    \centering
    \caption{}\label{fig:sequences}
    \includegraphics[width=1\textwidth]{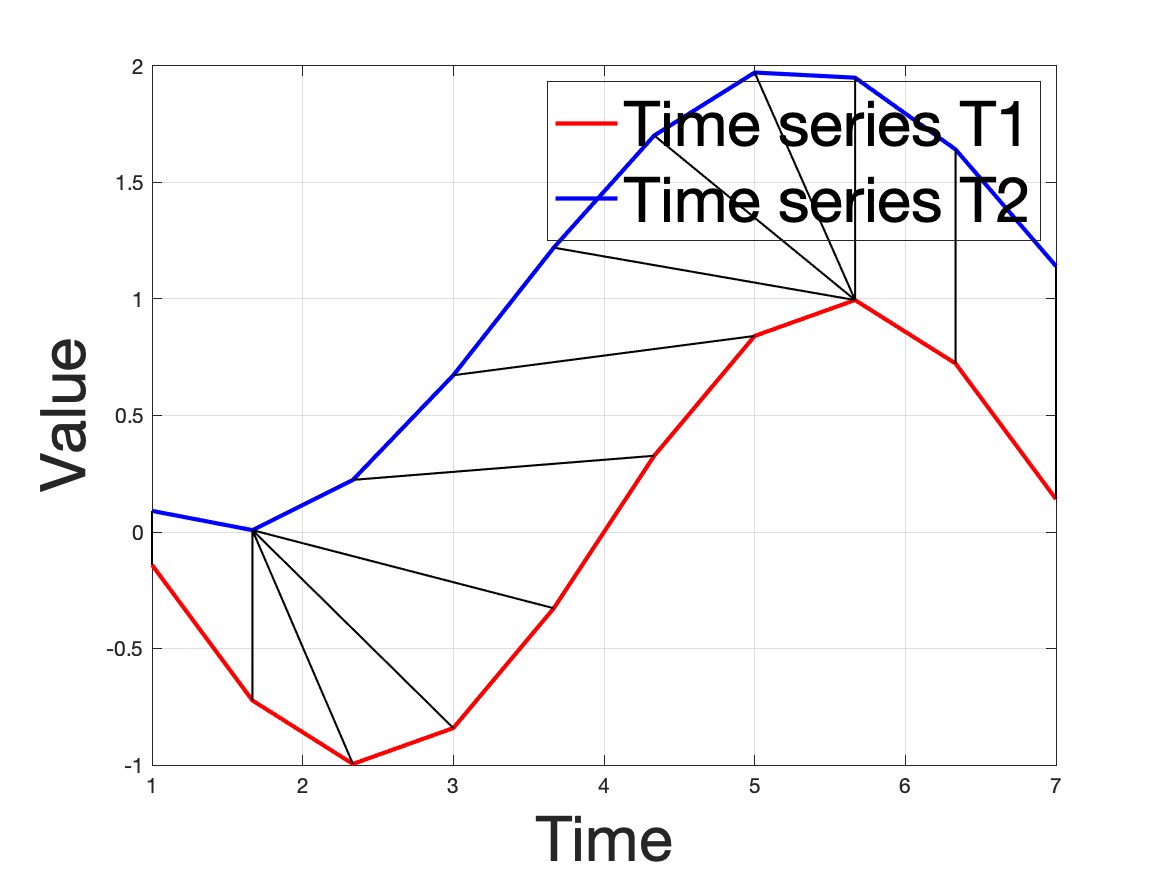}
\end{subfigure}
\caption{Illustration of finding an optimal alignment: (a) a matrix showing the warping path, i.e., an optimal alignment of time series $T1$ and $T2$ in DTW, where each cell $(i,j)$ is the distance between $ith$ element of $T1$ and $jth$ element of $T2$ and the DTW distance is the sum of distances on the optimal path shown in orange, and (b) optimal alignment of each point in time series T1 (blue) and time series T2 (orange) is shown with black lines.}
\label{fig:dtw_matrix}
\end{figure}

Aligning two time series $T1$ of length $a$ and $T2$ of length $b$ using DTW involves creating a $a \times b$ matrix, $D$, where the element $D[i,j]$ is the square of Euclidean distance, $d(t1_i,t2_j)^2$, where $t1_i$ is the $i$th point of $T1$, $t2_j$ is the $j$th point of $T2$, and $d(\cdot, \cdot)$ is the Euclidean distance. A warping path $P$ is a sequence of \cmmnt{contiguous} matrix elements that are mapped between $T1$ and $T2$ (Figure~\ref{fig:dtw_matrix}). This warping path must satisfy the following three conditions. 
\begin{itemize}
    \item Every point from $T1$ must be aligned with one or more points from $T2$, and vice versa.
    \item The first points of $T1$ and $T2$ must align, and so must their last points.  In other words, the warping path must start and finish at diagonally opposite corner cells of the matrix.
    \item No cross-alignment is allowed, that is, the warping path must increase monitonically in the matrix plot. %The mapping between points from $T1$ to $T2$ must increase monotonically, and vice versa. In other words, .
\end{itemize}

Among all paths in the matrix that satisfy the three conditions above, we are interested in finding the path that minimizes the distance calculated as shown in Equation~\ref{eq:dtw_min}~\citep{ShokoohiYekta2015}. 
\begin{equation}
\label{eq:dtw_min}
    \text{DTW}(T1,T2) = \min_{P \text{ between } T1 \text{ and } T2}{\sqrt{\sum_{(i,j)\equiv p_k \in P}{D[i,j]}}},
\end{equation}
\noindent which enumerates over all possible warping paths $P$ between $T1$ and $T2$ and finds the optimal warping path that minimizes the distance.
Algorithm~\ref{alg:DTW} outlines the steps to calculate DTW($T1$, $T2$). %, where $T1$ and $T2$ are two time series.
\begin{algorithm}[!ht]
\caption{Dynamic time warping algorithm for distance calculation between two \cmmnt{univariate} time series.}\label{alg:DTW}
\nonl \underline{Algorithm DTW}\\
\SetAlgoLined
\nonl Input: $T1$ and $T2$: time series, W: warping window size\\
\nonl Output: distance between $T1$ and $T2$
\smallskip\\
\nonl Procedure% ($Tweets$) 
\smallskip\\
Let $a$ and $b$ be the lengths of $T1$ and $T2$, respectively.\\
Let $m$ be the number of variables in $T1$ and $T2$ respectively.\\
Create a distance matrix $D$ of size $a \times b$ and initialize all matrix elements to $\infty$.\label{lin:DTW_matrix}\\
$D[0,0]$ := 0. // Initialize the first entry in $D$.\\
$i$ := 1. $j$ = 1. // Initialize the index of a warping path between $T1$ and $T2$.\\
\While{$i \leq a$ {\rm and} $j \leq b$}{
Calculate the squared Euclidean %%$d(t1_i,t2_j)^2$ 
distance, $\sum_{c=1}^{m}(t1_{i,c} - t2_{j,c})^2$, between the $i$th item in $T1$ and each of the $j$th item in $T2$ within the range of $j = [i-W, i+W]$.\label{lin:dist}\\
Update $D[i,j]$ to %distance($i,j$) 
$d(t1_i,t2_j)^2$ + $\min\{D[i-1,j], D[i,j-1], D[i-1,j-1]\}$.\label{lin:alg_dist}\\
increase $i$ by 1.\\
}
return $\sqrt{D[a,b]}$.
\end{algorithm}

In this work, the sensor times series are multivariate (precisely, bivariate)  defined by the discharge and SSC. We have considered two variants of DTW, DTW-independent (DTW-I) and DTW-dependent (DTW-D). DTW-I calculates the distance as the sum of distances that are calculated separately for individual variables (calling the DTW for each variable). %in an $m$-variate time series by treating the time series as $m$ separate univariate time series.   
DTW-D, on the other hand, handles $T1$ and $T2$ as multivariate time series and calls the DTW once. The dependency between discharge and concentration is important in our work and, therefore, we used DTW-D. % whenever dynamic time warping was employed to calculate distance between two storm event time series.
 %

%\byung{Notation may be confusing.}\ali{I have edited this section heavily in terms of names and short hand to try and build up to this point. I am unable to improve the notation of this particular equation to make it simpler.}
%\ali{replace the distance function in algorithm for DTW, with this equation without the looping over dimensions.}\ali{We mentioned here that we are using DTW-D because the dependency between multiple variables in our work can not be ignored and is what is of importance to us.}

\subsection{Experimental test cases}\label{sec:synth_data}
%There are two experimental test cases used for evaluation of the 3-dimensional clustering approach, (i) Synthetic dataset, (ii) Mad River dataset

\subsubsection{Synthetic hydrograph and sedigraph dataset}\label{sec:synthetic_dataset}

%\begin{wrapfigure}{r}{0.5\linewidth}
\begin{figure}[!ht]
\centering
\begin{subfigure}[b]{0.225\textwidth}
\centering
\caption{}\label{fig:synth_streamflowa}
\includegraphics[width=1\textwidth]{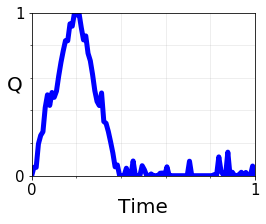}%\quad
\end{subfigure}
\begin{subfigure}[b]{0.225\textwidth}
\centering
\caption{}\label{fig:synth_streamflowb}
\includegraphics[width=1\textwidth]{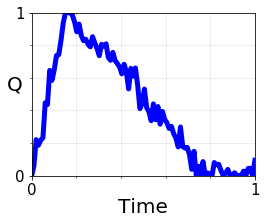}%\quad
\end{subfigure}
\begin{subfigure}[b]{0.225\textwidth}
\centering
\caption{}\label{fig:synth_streamflowc}
\includegraphics[width=1\textwidth]{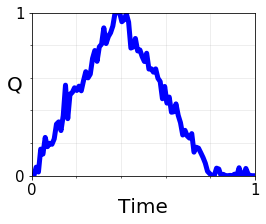}%\quad
\end{subfigure}
\begin{subfigure}[b]{0.225\textwidth}
\centering
\caption{}\label{fig:synth_streamflowd}
\includegraphics[width=1\textwidth]{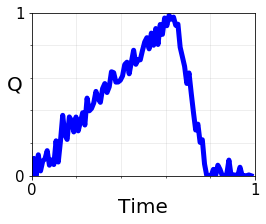}
\end{subfigure}
\\
\begin{subfigure}[b]{0.225\textwidth}
\centering
\caption{}\label{fig:synth_streamflowar}
\includegraphics[width=1\textwidth]{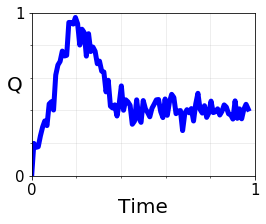}%\quad
\end{subfigure}
\begin{subfigure}[b]{0.225\textwidth}
\centering
\caption{}\label{fig:synth_streamflowbr}
\includegraphics[width=1\textwidth]{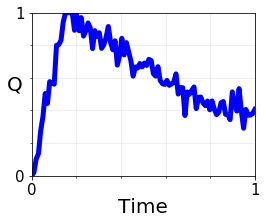}%\quad
\end{subfigure}
\begin{subfigure}[b]{0.225\textwidth}
\centering
\caption{}\label{fig:synth_streamflowcr}
\includegraphics[width=1\textwidth]{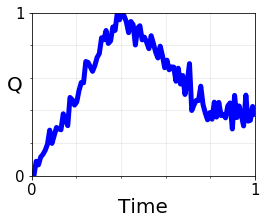}%\quad
\end{subfigure}
\begin{subfigure}[b]{0.225\textwidth}
\centering
\caption{}\label{fig:synth_streamflowdr}
\includegraphics[width=1\textwidth]{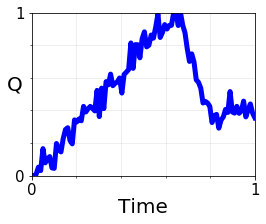}
\end{subfigure}
\\
\begin{subfigure}[b]{0.225\textwidth}
\centering
\caption{}\label{fig:synth_sediflowa}
\includegraphics[width=1\textwidth]{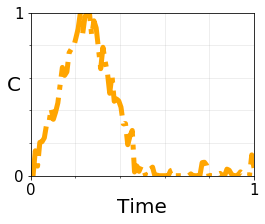}
\end{subfigure}
\begin{subfigure}[b]{0.225\textwidth}
\centering
\caption{}\label{fig:synth_sediflowb}
\includegraphics[width=1\textwidth]{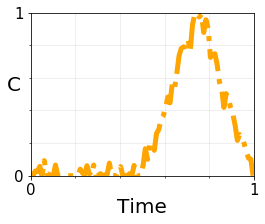}
\end{subfigure}
%\ali{Donna/Scott: For these synthetic events, should we be talking about them in generic sense of concentration or specifically as SSC. I remember earlier Scott and I decided to keep this to SSC since the trends are particular to SSC.}\donna{I keep forgetting to ask Scott whether we should refer to hydrographs and "sedigraphs" throughout. Also, the only ingredient that missing from your benchmark data set is that we did not capture whether the hydrograph drops back to baseflow conditions. I don't want to have this postpone submission of the manuscript. But after we submit, it might be a good idea to add this feature to the benchmark data as it might come up in review and we'd be ready. what do others think?}\ali{I agree, it will not be a lot of work to introduce 1-2 more classes that end with a higher base flow than they started with.}
\caption{Eight types of synthetic hydrographs: (a) Flashy -- very early peak -- complete recess, (b) Early peak -- complete recess, (c) Mid-peak -- complete recess, (d) Delayed peak -- complete recess, (e) Flashy -- very early peak -- incomplete recess, (f) Early peak -- incomplete recess, (g) Mid-peak -- incomplete recess, (h) Delayed peak -- incomplete recess,  and two different types of synthetic concentration graphs: (i) Early peak, (j) Late peak. }\label{fig:synth_graphs}
\end{figure}
%\end{wrapfigure}

To help validate the computational clustering method, we generated %large volumes of 
synthetic (i.e., not obtained through  measurement~\citep{syntheticData}) hydrographs and sedigraphs using domain knowledge. These synthetic data can %Synthetic datasets are popularly used in training machine learning models in what is known as transfer learning~\citep{Pan2010}. 
be labeled and used as the ground truth to facilitate the process of assessing methodology. A dataset generator was designed to produce synthetic hydrographs and concentration-graphs that contained realistic levels of sensor noise. 
Four control parameters were used: time-to-peak, duration-of-peak, onset, and recess. Time-to-peak controls the duration it takes for the concentration/discharge values to reach peak value of 1, duration-of-peak controls the duration of flow above base conditions, onset controls the ``time'' at which values (either concentration or discharge) start to rise in magnitude above base conditions, and recess controls the degree to which concentration/discharge values return to base conditions at the end of an event. Table~\ref{tbl:synth_param_setting} shows the values (between 0 and 1) for each of these parameters for the different types of synthetic graphs generated.

%Our goal was not to simulate precise representations of hydrograph and sedigraph shapes, but rather a controlled dataset that could be used for model validation. 
%Five control parameters were used: time-to-peak discharge and the duration-of-peak discharge %\donna{Should this be duration of discharge?}\ali{Let us discuss in next meeting. Duration of peak discharge is always 1 since when peak discharge ends, event ends. We played with this word to accomadate hydrograph type 1 where the "peak" discharge lasts only 0.3\% of time and then slowly gets back to zero but still stays above zero to keep event alive.}, 
%time-to-peak and duration of concentration, and the onset time of concentration. The onset of discharge is used to define the start of an event (i.e., time 0). Table~\ref{tbl:synth_param_setting} shows default values for each of the dataset parameters.%; for time-to-peak, a random value was selected from the range shown.

%\renewcommand\baselinestretch{1}\selectfont
\begin{table}[ht]
\caption{Default parameter settings for synthetic hydrograph and sedigraph generator.}\label{tbl:synth_param_setting}
\centering
\begin{tabular}{|c|c|c|c|c|}
\hline
\multicolumn{4}{|c|}{\bf Hydrograph}\\
\hline
{\bf Type} & {\bf Duration-of-peak}& {\bf Time-to-peak} &{\bf Onset} & {\bf Recess}\\
\hline
Flashy - very early peak - Complete Recess & 0.4 & 0.5 & 0 & 0\\
Flashy - very early peak - Incomplete Recess & 0.4 & 0.5 & 0 & 0.4\\
Early peak - Complete Recess & 0.8 & 0.2 & 0 & 0\\
Early peak - Incomplete Recess & 0.8 & 0.2 & 0 & 0.4\\
Mid-peak - Complete Recess & 0.8 & 0.5 & 0 & 0\\
Mid-peak - Incomplete Recess & 0.8 & 0.5 & 0 & 0.4\\
Delayed peak - Complete Recess & 0.8 & 0.8 & 0 & 0\\
Delayed peak - Incomplete Recess & 0.8 & 0.8 & 0 & 0.4 \\
\hline
\multicolumn{4}{|c|}{\bf Concentration-graph}\\
\hline
{\bf Type} & {\bf Duration}& {\bf Time-to-peak} & {\bf Onset} & {\bf Recess}\\
\hline
Early peak & 0.5 & 0.5  & 0 & 0 \\
Late peak & 0.5 & 0.5  & 0.5 & 0 \\

\hline
\end{tabular}

\end{table}

\begin{figure}[!ht]
\begin{subfigure}[b]{0.3\textwidth}
\centering
\caption{}
\includegraphics[width=1\textwidth]{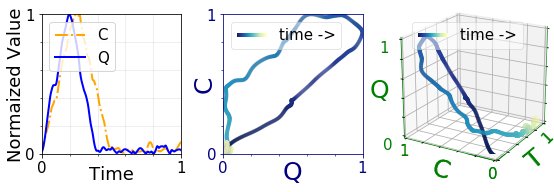}%
\end{subfigure}\qquad
\begin{subfigure}[b]{0.3\textwidth}
\centering
\caption{}
\includegraphics[width=1\textwidth]{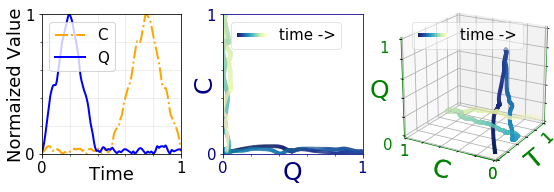}%
\end{subfigure}\qquad
\begin{subfigure}[b]{0.3\textwidth}
\centering
\caption{}
\includegraphics[width=1\textwidth]{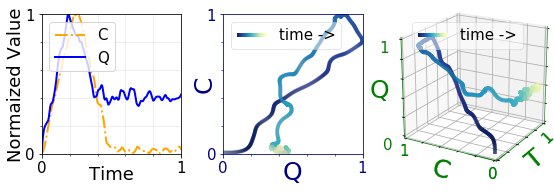}%
\end{subfigure}\\
\begin{subfigure}[b]{0.3\textwidth}
\centering
\caption{}
\includegraphics[width=1\textwidth]{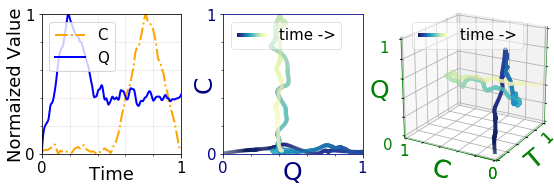}
\end{subfigure}\qquad
\begin{subfigure}[b]{0.3\textwidth}
\centering
\caption{}
\includegraphics[width=1\textwidth]{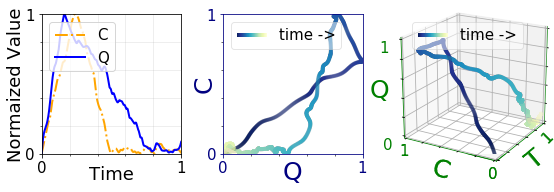}
\end{subfigure}\qquad
\begin{subfigure}[b]{0.3\textwidth}
\centering
\caption{}
\includegraphics[width=1\textwidth]{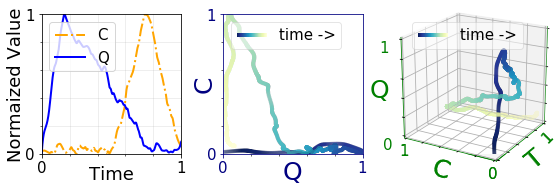}
\end{subfigure}\\
\begin{subfigure}[b]{0.3\textwidth}
\centering
\caption{}
\includegraphics[width=01\textwidth]{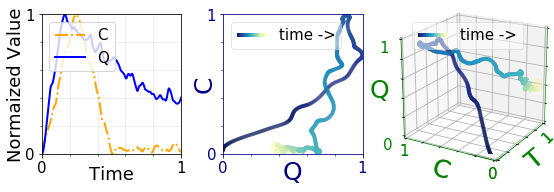}
\end{subfigure}\qquad
\begin{subfigure}[b]{0.3\textwidth}
\centering
\caption{}
\includegraphics[width=1\textwidth]{figures/synth_samples/synthetic_6_sample_0.png} %
\end{subfigure}\qquad
\begin{subfigure}[b]{0.3\textwidth}
\centering
\caption{}
\includegraphics[width=01\textwidth]{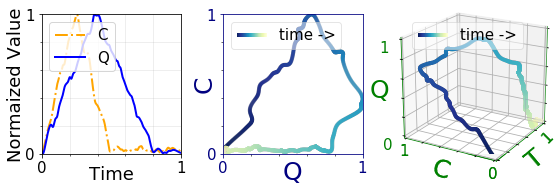}
\end{subfigure}\\
\begin{subfigure}[b]{0.3\textwidth}
\centering
\caption{}
\includegraphics[width=1\textwidth]{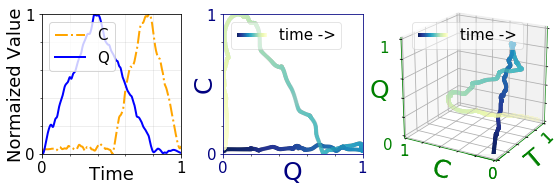} %
\end{subfigure}\qquad
\begin{subfigure}[b]{0.3\textwidth}
\centering
\caption{}
\includegraphics[width=01\textwidth]{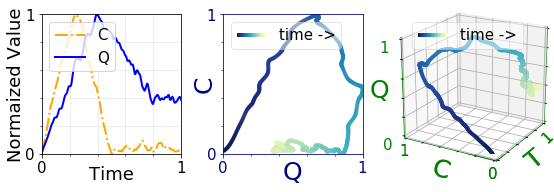}
\end{subfigure}\qquad
\begin{subfigure}[b]{0.3\textwidth}
\centering
\caption{}
\includegraphics[width=1\textwidth]{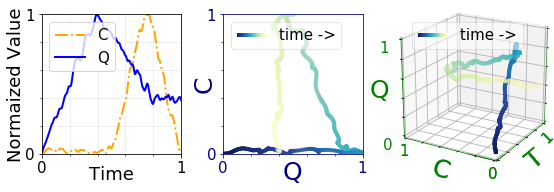} %
\end{subfigure}\\
\begin{subfigure}[b]{0.3\textwidth}
\centering
\caption{}
\includegraphics[width=01\textwidth]{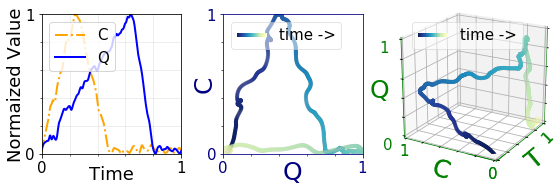}
\end{subfigure}\qquad
\begin{subfigure}[b]{0.3\textwidth}
\centering
\caption{}
\includegraphics[width=1\textwidth]{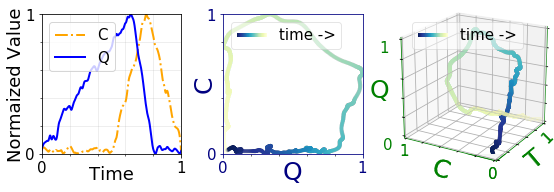} %
\end{subfigure}\qquad
\begin{subfigure}[b]{0.3\textwidth}
\centering
\caption{}
\includegraphics[width=01\textwidth]{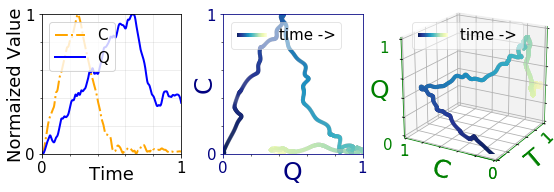}
\end{subfigure}\\
\centering
\begin{subfigure}[b]{0.3\textwidth}
\centering
\caption{}
\includegraphics[width=1\textwidth]{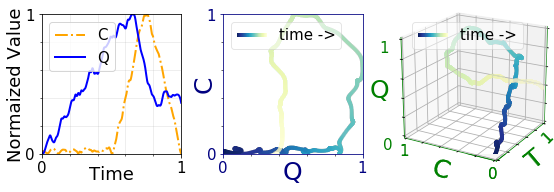} %
\end{subfigure}

\vspace*{-2.5ex}
\caption{An example event from each class in the synthetic dataset: (a) Class 1, (b) Class 2, (c) Class 3, (d) Class 4, (e) Class 5, (f) Class 6, (g) Class 7, (h) Class 8, (i) Class 9, (j) Class 10, (k) Class 11, (l) Class 12, (m) Class 13, (n) Class 14, (o) Class 15, and (p) Class 16. }\label{fig:synthetic_trajectories}
\end{figure}

Examples of eight types of synthetic hydrographs and two types of synthetic concentration-graphs (see Figure~\ref{fig:synth_graphs}) %\ali{What will be a generic name for sedigraphs? concentration-graphs?} 
shows different timings for the rise and fall of discharge and the concentration, respectively.
Thus, when combining the eight types of hydrographs and the two types of concentration-graphs, the synthetic data represent sixteen possible types of storm events (see Figure~\ref{fig:synthetic_trajectories}). 
Random samples from a normal (Gaussian) distribution with a mean of $0.00$ and standard deviation of $0.05$ were added to the discharge and concentration value at each time step. We generated 800 synthetic storm events this way, equally distributed among the sixteen types.

%Uniform noise from the half-open interval $(-0.1, 0.1]$ to the default value was added to the discharge and SCC at each time-step. We generated 8,000 synthetic storm events this way, equally distributed among the eight types.

%\subsection{Synthetic dataset}\label{apx:synth_trajectories}

%\renewcommand\baselinestretch{1}\selectfont

\subsubsection{Application to real hydrograph and sedigraph dataset}\label{sec:real_dataset}

We applied the event clustering method to the \cmmnt{primary} Mad River watershed data. The time series of discharge and SSC were pre-processed (see Section~\ref{sec:preprocess}), and used as input to K-medoids with the DTW-D algorithm. The resulting clusters of events were examined with respect to the following: (i) optimal number clusters, (ii) relationship to hysteresis loop (iii) relationship to watershed sites, (iv) discrimination of event characteristics through clustering and hysteresis, and (v) characteristic of event clusters (based on storm event metrics).
%, and (vi) variation in categories in relation to number and types of watersheds.%\donna{We either need to elaborate upon these here as I don't think they'll have meaning to the reader at this point in the story. It might be easier to remove this here and refer to something like this in the discussion after you've explained the analysis to the reader}\ali{This was originally the preamble of the results section. It is meant to give readers a headsup of what is to come. We can remove this section if it is not needed here, it probably wont make much sense in discussion after we have talked about all this in detail.}

\subsubsection{Finding the optimal number of clusters on the real data }\label{sec:elbow}
In this work, we identified the ``optimal'' number of clusters for classification using the elbow method, in which the sum of squared errors (SSEs) are plotted against an increasing number of $K$ clusters. A value for $K$ is selected (visually) as the point where further increases in $K$ result in diminishing reduction in SSE, thus making the onset of the plateau look like an elbow of an arm. In this respect, optimal $K$ values were selected empirically, and then further validated using the Kneedle algorithm~\citep{Satopaa2011}). 

\subsection{Evaluation criteria}\label{sec:accuracy_measures}

\subsubsection{Clustering measures}

\begin{figure}[!htb]
\centering
\includegraphics[width=0.6\textwidth]{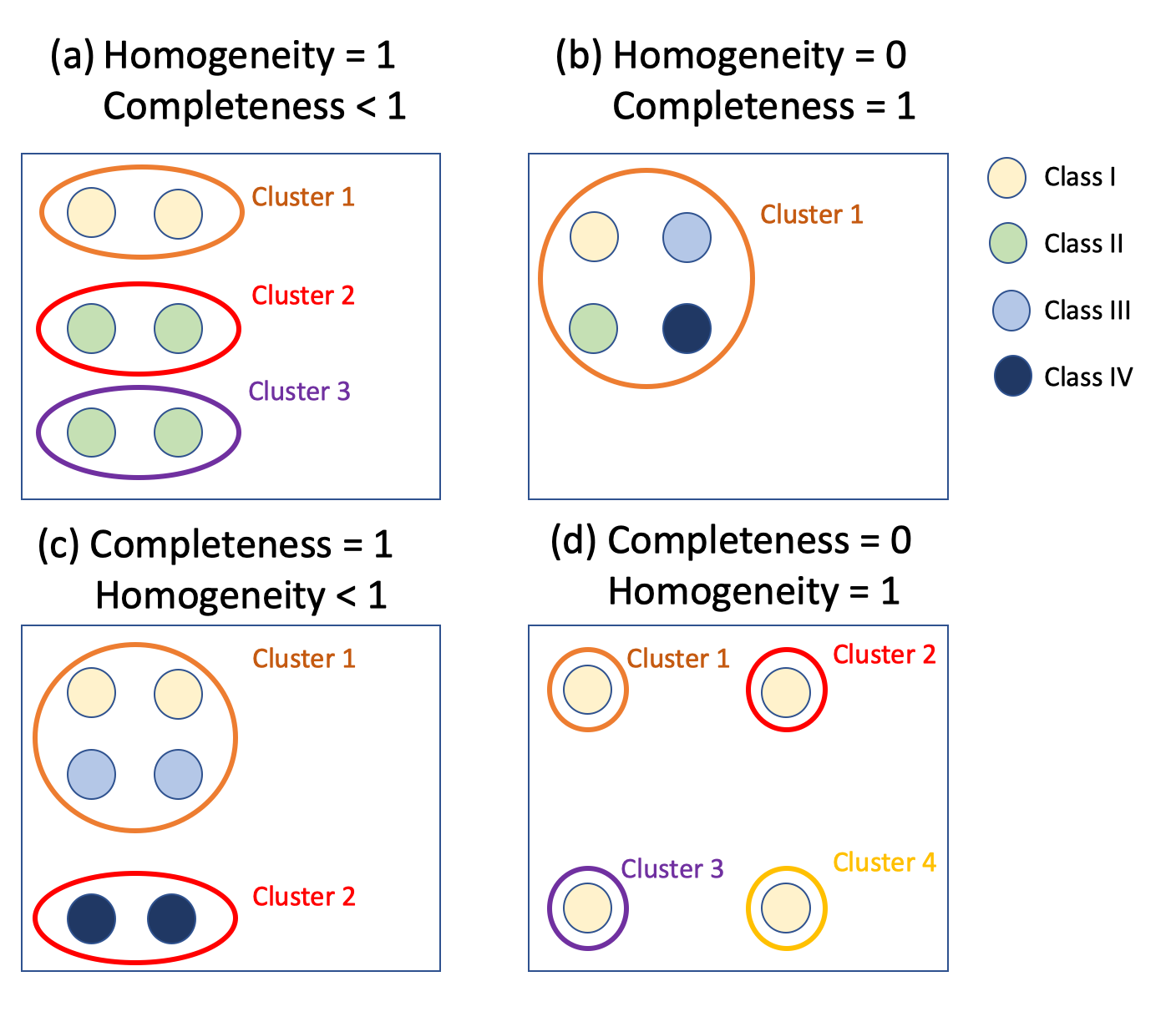}
\caption{Schematic illustration of the measures -- completeness and homogeneity with class representing ground-truth classifications.}\label{fig:evaluation_examples}
\end{figure}

We used three external measures of clustering to evaluate K-medoids with DTW-D on the synthetic storm event dataset --- (a) Rand Index (b) Homogeneity and (c) Completeness. \textit{Rand Index} is perhaps the most commonly used similarity measure between two different partitionings of a dataset, and is defined as the ratio of correct decisions over all decisions made, where a decision is made for each pair of elements in regard to putting both elements of the pair in the same cluster or different clusters. 
%number of pairs of elements that are either in the same group or in different groups in both parititonings over the number of all pairs of elements. 
%\donna{I think it might be easier for our target audience to think of the Rand index in terms of true positives and true negatives. What do you think? We could then explain it as the percentage of "correct decisions" made by the algorithm. This also would avoid our referencing "two" groups as things are either correctly classified or incorrectly classified.} 
Its value ranges 0.0 to 1.0, where 1.0 means that the groups are identical and 0.0 means that the two partitionings do not agree on any pair of elements. %Equation~\ref{eq:ri} is used to calculate Rand Index. 
    %\begin{equation}
    %    Rand Index = \frac{a+b}{{n \choose 2}}
    %\end{equation}\label{eq:ri}
    %\noindent where n is the total number of objects, a is the number of pairs of objects in the same class in both partitions of data and b is the number of pairs of objects in different classes in both partitions of data. 
 %
\textit{Homogeneity} ranges from 0.0 to 1.0, where 1.0 means that every cluster contains only elements that are members of the same class (see Figure~\ref{fig:evaluation_examples}a) and 0.0 means that there is only one cluster and every element in it belongs to a different class (see Figure~\ref{fig:evaluation_examples}b)~\citep{Rosenberg2017}. 

%\begin{wrapfigure}{r}{0.40\linewidth}
%\end{wrapfigure}

\textit{Completeness} ranges from 0.0 to 1.0, where 1.0 means that all elements of any given class are in the same cluster (see Figure~\ref{fig:evaluation_examples}c) and 0.0 means that there is only one class and every element in it is assigned to a different cluster 
(see Figure~\ref{fig:evaluation_examples}d)~\citep{Rosenberg2017}.

\subsubsection{Statistical measures}
%
%\donna{I also think we may want to remove all of this section...no? I know we started the discussion as to what these statistical measures are really telling us. I'mstill having a tough time getting my head around what an ANOVA  and t-test are telling us in this context. It may be important to leave some of this here, but you and Scott are going to have to explain this. I really think that clustering algorithms should best be described in terms of the metrics you are using above (perhaps in addition to confusion matrices); but let's talk about this as a group.}\ali{Acknowledged. Just for my own note taking: The intent of ANVOA was to try and bring in a non subjective test, when we describe them in terms of metrics it is very subjective based on interpretation. If it is not the appropriate test to use here, lets not use it.}
We used four statistical measures to study clustering performance -- %(a) Two-sided t-test, 
(a) Chi-squared test of independence, (b) One-way analysis of variance (ANOVA), (c) Z-score, and (d) Hopkins test.
%
%To compare the accuracy of different clustering algorithms \chgd{on the UCR repository datasets} we used a two-sided t-test for similar means on ten runs of each algorithm. A \textit{two-sided t-test} tests the null hypothesis that two independent samples have similar means. A p-value of less than 0.05 was considered a statistically significant. 
%
To investigate the dependency between computational clusters and watershed sites/hysteresis loop categories we used a Chi-squared test of independence. This test is used to determine if there is a significant relationship between two categorical variables. The null hypothesis for the test is that there is no relationship between the variables.
We performed ANOVA test to investigate how well the 24 storm event metrics (see Section~\ref{sec:studyArea}) for the storm events are explained by the Wiliams's hysteresis loop classes and the computational clusters, respectively. \textit{ANOVA} tests the null hypothesis that all groups have the same population mean. It does so by calculating the f-value as the ratio of the variance among group means over the average variance within groups to determine the ratio of explained variance to unexplained variance. For our purpose of clustering, a larger f-value indicates more accurate clustering, i.e, with higher inter-group similarity and lower intra-group similarity. 
%We do not use ANOVA test to investigate how well the storm event metrics are explained for Hamshaw's types because of low number of events in each type.\ali{Do we need this last line? I do not think so. }
%
To analyze the characteristics of each cluster using the 24 storm event metrics we used z-score value of each metric for each cluster. \textit{Z-score} measures the number of standard-deviations the value of a metric is different from the mean metric value of all storm events. We compared the average metric value of events within a cluster to the average metric value of all events in the dataset.
To measure the cluster tendency of a data set, we used Hopkins Test~\citep{Banerjee_2004}. \textit{Hopkins test} tests the null hypothesis that data are generated by a Poisson point process and thus are uniformly randomly distributed. A value close to 1 indicates the data is highly clusterable, while a value close 0 indicates the data uniformly distributed.

\section{Results}\label{sec:experiments}

\subsection{Validation of method using the synthetic dataset}

\begin{wrapfigure}[13]{l}{0.4\linewidth}
%\begin{figure}%[ht]
\centering
\includegraphics[width=0.33\textwidth]{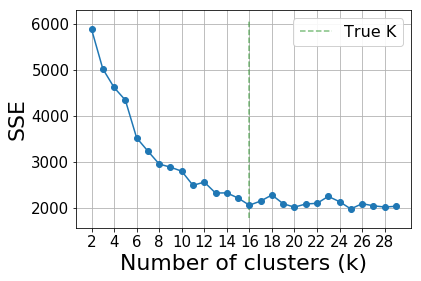}
\caption{Sum of square error (SSE) for different number of clusters from the synthetic storm event dataset. (True value of  $k$=16.)}\label{fig:best_k_av_synth}
%\end{figure}
\end{wrapfigure}

Clusters resulting from K-medoids with DTW-D were identical to the ground truth (see Section\ref{sec:synth_data}). That is, K-medoids with DTW-D showed the score of 1.0 for all Rand Index, homogeneity and completeness despite the noise inherent in the synthetic dataset. 
Additionally, the synthetic dataset had a Hopkins test statistic of 1.00 indicating its high clusterability and suitability for use in clustering methods. 
Further, the elbow plot (see Figure~\ref{fig:best_k_av_synth}) \cmmnt{for synthetic data} showed the elbow to be \cmmnt{approximately close to the real value of} $k=16$, after which the reduction in SSE was negligible. 
These results thus confirmed the validity of the method used.
\begin{comment}
The performance of K-medoids with DTW-D %was validated using synthetic dataset and 
showed promising results for accuracy in terms of Rand index, homogeneity, and completeness. This accuracy was expected from the validated high clusterability of the the synthetic dataset.% was highly clusterable, i.e., not uniformly distributed. 
Finally, the efficacy of using the elbow plot was also validated by observing an elbow close to the true $k$ value of synethtic dataset.
\end{comment}

%For our purpose of categorizing storm events, the dependency between the two variables, discharge and SSC, is an important factor in defining types of hydrological events (see Section~\ref{sec:hysteresis_loops}) and, therefore, we used DTW-D (with the dimensionality $M$ = 2) over DTW-I for multivariate time series in our work (Section~\ref{sec:dtw}).
%The marginal classification error was due to the noise injected into the dataset, which caused some instances of Class 5 to be mis-classified into Class 7 and some instances of Class 6  into Class 8 (see Table~\ref{tbl:synth_distribution}). Indeed, Figure~\ref{fig:synthetic_trajectories} in Section~\ref{sec:synthetic_dataset} shows some similarities between Class 5 (Figure~\ref{fig:synthetic_trajectories}e and Class 7 (Figure~\ref{fig:synthetic_trajectories}g) and between Class 6 (Figure~\ref{fig:synthetic_trajectories}f) and Class 8 (Figure~\ref{fig:synthetic_trajectories}h).

Note that K-medoids with DTW-D was chosen not only for its high accuracy but also for the following advantages over each of the three other selected algorithms (see Section~\ref{sec:multi-clustering}). TADPole requires an additional input parameter that is not very intuitive, in addition to the number of clusters (i.e., $K$). K-shape is inherently a univariate time series clustering algorithm and not applicable to multivariate time series. K-medoids with Euclidean distance does not have the flexibility of time series warping present in K-medoids with DTW-D, a quality we need in our algorithm given the approximate nature of event segmentation.

%\ali{to be added in discussion that even though we got 100\% accuracy in synthetic dataset, we do not expect 100\% accuracy in real dataset since real dataset is more continuous in nature than synthetic dataset which has is well seperated and has perfect clusterability.}
%%%%%%%%%%%%%%%%%%%%%%%%%%%%%%%%%%%%%%%%%%%%%%%%%

\begin{comment}
%\renewcommand\baselinestretch{1}\selectfont
%\begin{wraptable}{r}{0.4\linewidth}
\begin{table}%[ht]
\caption{Accuracy measures for K-Medoids with DTW-D on the synthetic benchamark dataset.}\ali{is this table needed}\label{tbl:synth_results}
\centering
\begin{tabular}{|c|c|}
\hline
{\bf Accuracy measure} & {\bf Score} \\
\hline
Rand index &1.00\\
Completeness  &1.00\\
Homogeneity  &1.00\\
\hline
\end{tabular}
\end{table}
%\end{wraptable}
%\renewcommand\baselinestretch{2}\selectfont

\end{comment}
%%%%%%%%%%%%%%%%%%%%%%%%%%%%%%%%%%%%%%%%%%%%

\subsection{Application to real hydrograph and sedigraph dataset}\label{sec:app_to_real_data}

\subsubsection{Optimal number of clusters}
%\scott{Try re-wording to start with what we found (we identified 4 clusters of events in the 603 storm events. Then can explain further.}\scott{I would add sentence or two reporting the number of events in the clusters. Don't need to say each one, but could give range, highest and lowest, etc. You could also give a little description of each cluster - for example, cluster 1 events tend to have early, short duration peak of SSC, fast rise of hydrograph, and nearly full recession of streamflow back to baseflow, broad clockwise patterns. Cluster 2 - broader sedigraph, hydrograph recession not as complete; Cluster 3 - very early and brief SSC peak, multi-peaked sedigraphs; Cluster 4 - Delayed rise of hydrograph and sedigraph.}

\begin{wrapfigure}{r}{0.4\linewidth}
%\begin{figure}%[ht]
\centering
\includegraphics[width=0.33\textwidth]{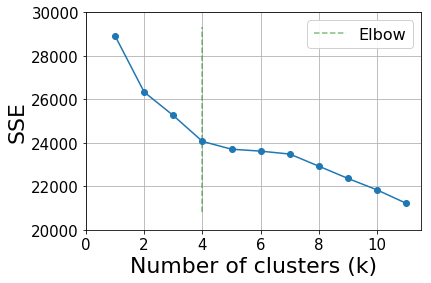}
\caption{Sum of square error (SSE) for different number of clusters from the \cmmnt{primary} Mad River storm event dataset. (The elbow is at $k$=4.)}\label{fig:best_k_av}
%\end{figure}
\end{wrapfigure}

\begin{figure}[!htp]
\begin{framed}
\begin{subfigure}{\textwidth}
\centering
\caption{Cluster 1. n = 234 }\label{fig:av_metadata_cluster_k4_1}
\includegraphics[width=0.3\textwidth]{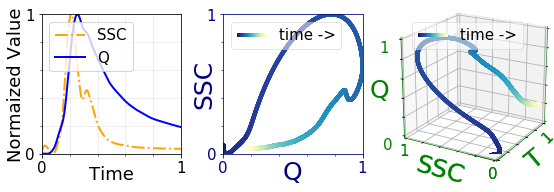}\qquad 
\includegraphics[width=0.3\textwidth]{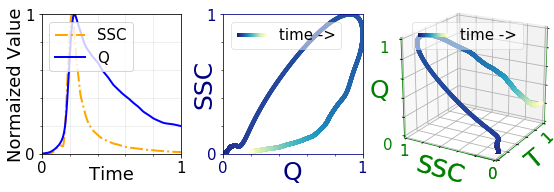}\qquad 
\includegraphics[width=0.3\textwidth]{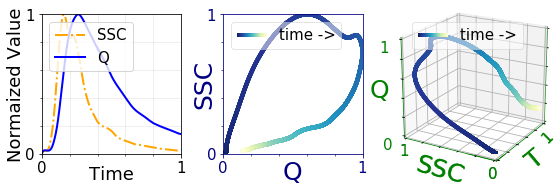}\\ 
\includegraphics[width=0.3\textwidth]{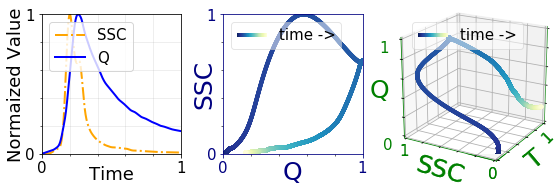}\qquad
\includegraphics[width=0.3\textwidth]{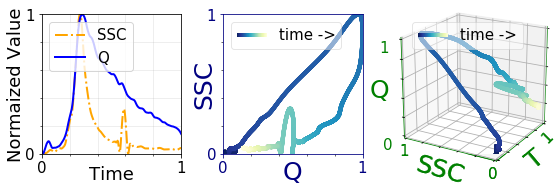}\qquad 
\includegraphics[width=0.3\textwidth]{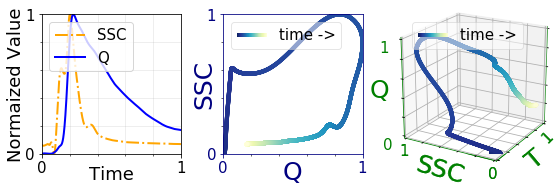}
\end{subfigure}
\end{framed}
\begin{framed}
\begin{subfigure}{\textwidth}
\centering
\caption{Cluster 2. n = 125}\label{fig:av_metadata_cluster_k4_2}
\includegraphics[width=0.3\textwidth]{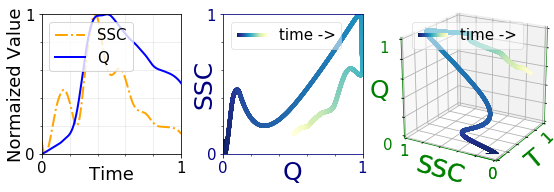}\qquad 
\includegraphics[width=0.3\textwidth]{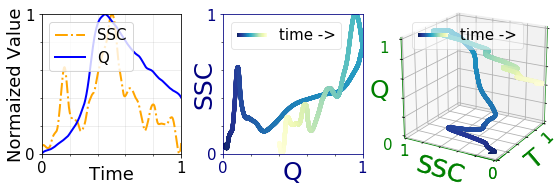}\qquad 
\includegraphics[width=0.3\textwidth]{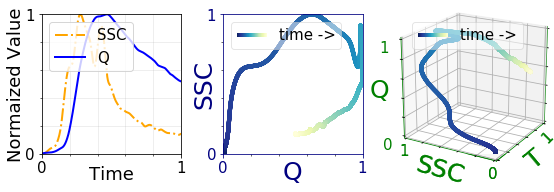}\\ 
\includegraphics[width=0.3\textwidth]{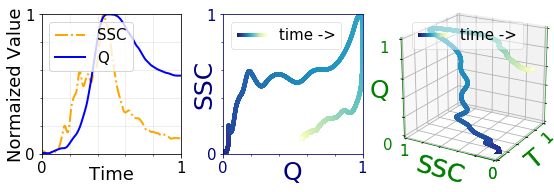}\qquad
\includegraphics[width=0.3\textwidth]{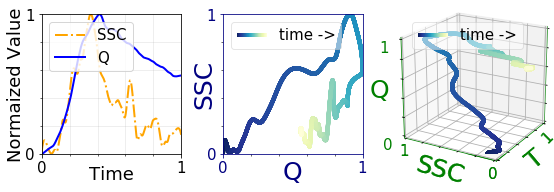}\qquad 
\includegraphics[width=0.3\textwidth]{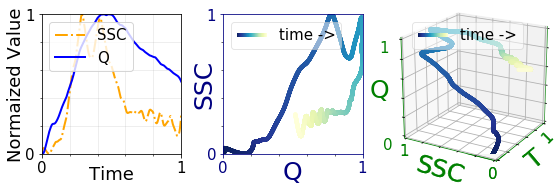}
%
%\vspace*{-2.5ex}

\end{subfigure}
\end{framed}
\begin{framed}
\begin{subfigure}{\textwidth}
\centering
\caption{Cluster 3. n = 116}\label{fig:av_metadata_cluster_k4_3}
\includegraphics[width=0.3\textwidth]{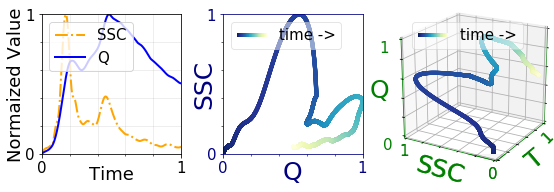}\qquad 
\includegraphics[width=0.3\textwidth]{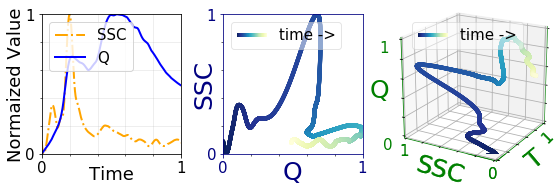}\qquad 
\includegraphics[width=0.3\textwidth]{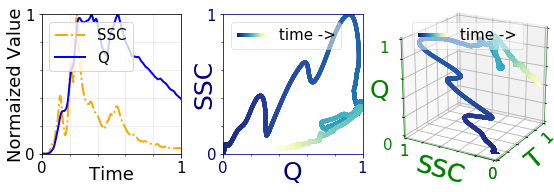}\\
\includegraphics[width=0.3\textwidth]{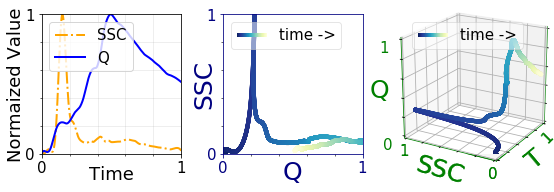}\qquad
\includegraphics[width=0.3\textwidth]{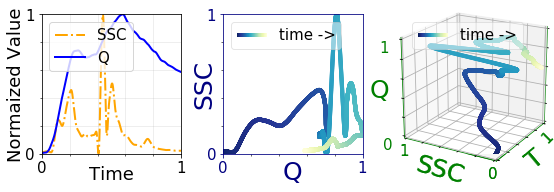}\qquad 
\includegraphics[width=0.3\textwidth]{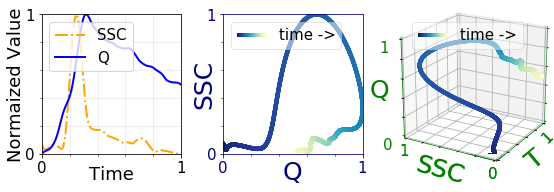}
%\vspace*{-2.5ex}

\end{subfigure}
\end{framed}
\begin{framed}
\begin{subfigure}{\textwidth}
\centering
\caption{Cluster 4. n = 128}\label{fig:av_metadata_cluster_k_4}
\includegraphics[width=0.3\textwidth]{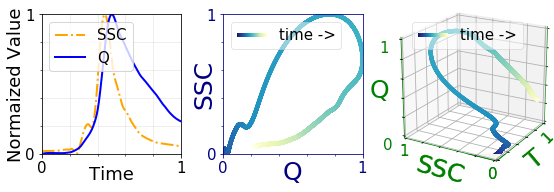}\qquad
\includegraphics[width=0.3\textwidth]{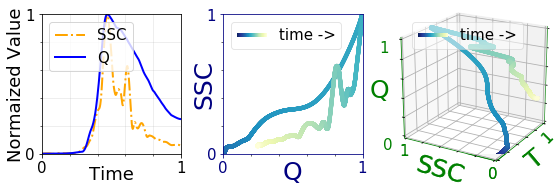}\qquad
\includegraphics[width=0.3\textwidth]{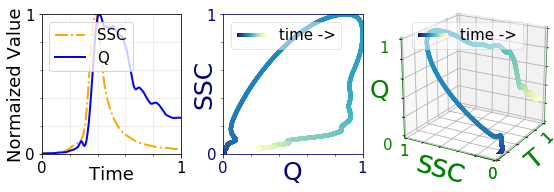}\\
\includegraphics[width=0.3\textwidth]{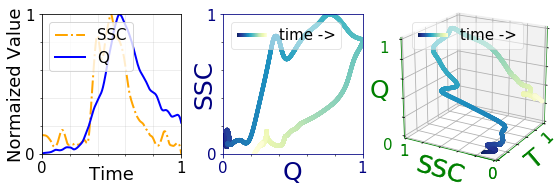}\qquad
\includegraphics[width=0.3\textwidth]{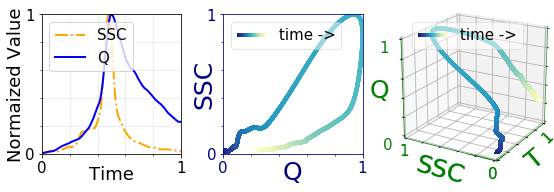}\qquad
\includegraphics[width=0.3\textwidth]{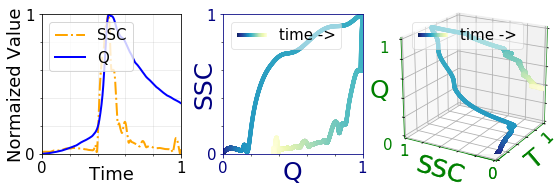}
%\vspace*{-2.5ex}

\end{subfigure}
\end{framed}
%\vspace*{-3ex}
\caption{Six storm events closest to the centroid in each cluster of \cmmnt{primary} Mad River dataset ($k=4$, $N = 603$).}\label{fig:av_metadata_cluster_k4}
\end{figure}

The \cmmnt{primary} Mad River dataset had a Hopkins test statistic of 0.96, which indicates that the dataset is highly clusterable and, therefore, suitable to be used in clustering methods. Application of the K-medoids with DTW-D to the \cmmnt{primary} dataset (N = 603 storm events) yielded $k=4$ clusters using both the the elbow technique (Figure~\ref{fig:best_k_av}) and the Kneedle algorithm~\citep{Satopaa2011} (Section~\ref{sec:elbow}).
The sizes of clusters were approximately 120 for three of them and 234 for one of them. %\donna{I'm not certain if this previous sentence is needed anymore? Was it referring to the n values associated with the clusters of Figure 10? Also, if we're going to use "N" number of data in sub-clusters, then let's use "n" for the total value. Note: I usually reverse the notation.}\ali{Yes, that is correct. Lets use n for number of points in cluster, and N for total number of points. I usually do that to.}
Cluster 1 events tended to have broad clockwise hysteresis patterns with an early, and relatively short peak duration for SSC; the hydrograph raised quickly and nearly fully returned to baseflow (see Figure~\ref{fig:av_metadata_cluster_k4_1}). %\donna{Please make certain that I haven't changed the intent throughout this section. I think all of us need to talk about how to deal with the direction of the hydrographs. It's important to compare algorithms as fairly as possible; but also show the algorithm advantages (e.g., in this case hysteresis could be first pre-processed by direction).}
Cluster 2 events tended to have broader (less flashy) sedigraphs and hydrographs with streamflow not returning completely to baseflow levels
(see Figure~\ref{fig:av_metadata_cluster_k4_2}). %\donna{The discussion should highlight the importance of this for both cluster 2 and 3 as this is likely indicative of saturated soils, high groundwater tables or soil moisture. Was there a predominant direction for the hysteresis loops?}\scott{agreed, the discussion should connect to antecedent conditions as well as groundwater flows}
Cluster 3 events were similar to cluster 2, but had flashier and often multi-peaked sedigraphs that were shorter in duration (see Figure~\ref{fig:av_metadata_cluster_k4_3}). %\scott{Ali, what did you mean by linear clockwise?}\ali{Thanks for catching that, I removed the sentence, the intent was to say that cluster 3 is dominated by Type 2-D and Type 4, i.e. linear than clockwise, however it is too much detail at this point and is discussed in next section.} \donna{Is there a way to backout a better description of what cluster 3 captures based on the z-scores of Figure 14?}\ali{The zscore plots are quiet different, attempting a better description, maybe Scott can help?}\scott{I think the basic descriptions of characteristics of hydrograph/sedigraph are ok - to me the zscore plots should come in later. Here we are just looking at what we can see in the data we are clustering, namely the sedigraph and hydrograph} 
The timing of the peak of the sedigraph and hydrographs of cluster 4 events were typically delayed and tended to have an initial period of slow rise of sedigraph and hydrograph prior to the period of rapid rise (see Figure~\ref{fig:av_metadata_cluster_k_4}). Events in cluster 4 tended to have hydrographs that return to near baseflow levels in contrast to cluster 2 and 3 events.
%\donna{Again, I'm not certain that I see this? I think the real difference between cluster 2 and 4 is that the sedigraph returns to a baseline of sorts...no?}\ali{discussed with Donna and fixed the description of cluster 4.}

%\subsection{Clusters from real data set}\label{apx:real_trajectories}
%Figure~\ref{fig:av_metadata_cluster_k4} shows sample trajectories of storm events in each of the four clusters resulting from the real storm event datasets. The samples are the top four trajectories closest to the centroid in the cluster. \byung{Let me make three comments on these four clusters: (1) there is quite decent coherency per cluster, which is good; (2) cluster 1 and cluster 2 look more similar than the others, and cluster 3 and cluster 4 look more similar than the others; how about checking the locations of their centroids to confirm that indeed the centroids of cluster 1 and cluster 2 are closer to each other than the others and the centroids of cluster 3 and cluster 4 are closer to each other than the others?; (3) it is hard for me to tell a story distinguishing between cluster 1 and cluster 2, nor between cluster 3 and cluster 4; there are some visible consistent differences in the shapes of stream flow and SSC, but what do these differences mean in terms of what's happening at the watersheds?}

%\renewcommand\baselinestretch{1}\selectfont

%\renewcommand\baselinestretch{2}\selectfont

%\begin{figure}\ContinuedFloat
%\begin{subfigure}{\textwidth}
%\centering
%\includegraphics[width=0.3\textwidth]{figures/av_metadata/samples/av_metadata_k4_Cluster_6_sample_0.png}\qquad %% 
%\includegraphics[width=0.3\textwidth]{figures/av_metadata/samples/av_metadata_k4_Cluster_6_sample_1.png}\\ %% 
%\includegraphics[width=0.3\textwidth]{figures/av_metadata/samples/av_metadata_k4_Cluster_6_sample_2.png}\qquad %%
%\includegraphics[width=0.3\textwidth]{figures/av_metadata/samples/av_metadata_k4_Cluster_6_sample_3.png}
%\caption{Cluster 6.}\label{fig:av_metadata_cluster_k9_6}
%\end{subfigure}

%\caption{Four trajectories closest to the centroid in each cluster of real trajectories ($k=9$).}\label{fig:av_metadata_cluster_k9}
%\end{figure} %% Figure file

\subsubsection{Relationship to hysteresis loops}\label{sec:rel_to_hyst}

%\scott{I'm note sure this paragraph adds much to the results section, I would drop. Could mention this in the discussion where talk about identifying number of clusters.}Table~\ref{tbl:scottDist} shows how the events from the Mad River dataset are distributed over Williams' six classes and Hamshaw's fifteen types. Interestingly the number of clusters, four, is similar to the number of Williams' six classes. Thus, further comparison was done against Williams' six classes. Note that since Hamshaw's 15 classes are refinement of Williams's six classes, any relationship present between Hamshaw's fifteen classes and computational clusters should also be observed with Williams' six classes. 
%\donna{I agree with Scott's comment above. In fact, I'm not certain we want to get into a comparison of the this method with Scott's. Since Williams' classification is more "accepted" it might be best to simply compare this method to that one? Let's think about the value added by this method so as to leverage using this method in tandem with either Williams or Scott's.}

Event cluster assignments did not correspond directly to
the C-Q hysteresis classifications  (see Section~\ref{sec:hysteresis_loops})
%the categorization of events based on their 2-D hysteresis patterns (Williams' six classes see Section~\ref{sec:hysteresis_loops}). 
The Mad River watershed events are severely skewed in the distribution of their hysteresis patterns when classified using Williams' classes (Table~\ref{tbl:scottDist}). That is, 63.8\% of the 603 events belong to Class II and each of the remaining five classes contains only 5--10\% of the events. In contrast, our computational clusters were relatively better balanced, with the largest of the four clusters containing only 39\% events and all of the remaining clusters containing more than 18\% of all events. Given the similarity of all the Mad River watershed sites (i.e., predominantly forested, mountainous watersheds), we would expect hydrological events to show a degree of similarity. This expectation was confirmed in the preponderance of Class II (clockwise) hysteresis patterns in the data. However, we found that our clustering did not classify events skewed to a single cluster, which would be expected given that our clustering method emphasizes the temporal aspect of the hydrograph and sedigraph \cmmnt{so much} more than the hysteretic aspect. %classification.
%

%\begin{wrapfigure}{r}{0.6\linewidth}
\begin{figure}
%\begin{figure}[!ht]
\centering
\begin{subfigure}[t]{0.45\textwidth}
\centering
\includegraphics[width=1\textwidth]{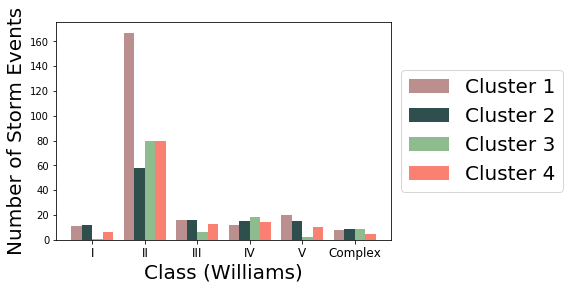}
\caption{Number of storm events in each class.}
\end{subfigure}%
\quad
\begin{subfigure}[t]{0.45\textwidth}
\centering
\includegraphics[width=1\textwidth]{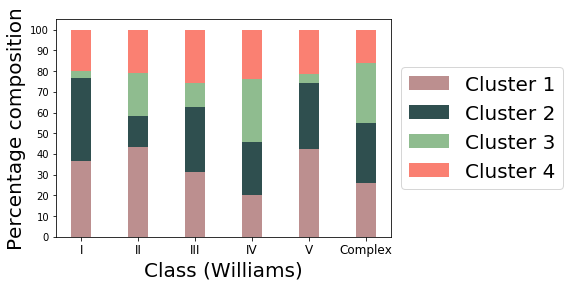}
\caption{Percentage of storm events in each class.}
\end{subfigure}\\

\begin{subfigure}[t]{0.45\textwidth}
\centering
\includegraphics[width=1\textwidth]{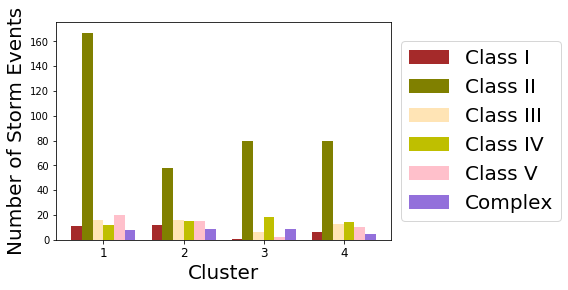}
\caption{Number of storm events in each cluster.}\label{fig:will_distributionc}
\end{subfigure}%
\quad
\begin{subfigure}[t]{0.45\textwidth}
\centering
\includegraphics[width=1\textwidth]{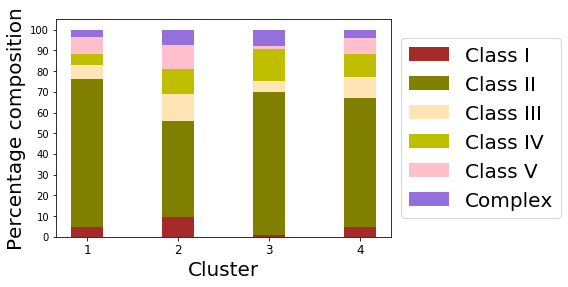}
\caption{Percentage of storm events in each cluster.}\label{fig:will_distributiond}
\end{subfigure}
\caption{Distribution of Williams' six classes in each cluster.}\label{fig:will_distribution}
\end{figure}
%\end{wrapfigure}

%%[Correlation]d
%\scott{I think it would be worthwhile to add a couple sentences describing how we see trends related to certain "Hamshaw" classes predominantly being in one cluster (e.g. Type 2-D in 3 and 2-B, 5-A in 1). While comparison to the "Hamshaw" classification is not focus of paper, it supports this point that hysteresis classification is not wholly independent of clustering.} 

%\donna{I like showing the distribution of your classification with Williams'. The fact that you're method shows fairly even distribution of Williams' classification into your clusters means that your method is identifying a completely separate set of characteristics. The key will be identifying these and articulating it. Note: I think we meed to also show your clustering into Williams classifications...maybe all as one figure (4 panels).}
On the other hand, %our computational study 
statistical test revealed that hysteresis classes and computational clusters were not independent of each other, hence correlated. Specifically, Chi-squared test of independence between hysteresis loop classes (both Williams' classes and Hamshaw's types) and computational clusters had a p-value lower than 0.001, thus establishing that they are not independent.
For instance, (i) 43\% of events in Class II appeared in the cluster 1 compared with no more than 21\% in each of the other clusters, and (ii) only one (out of 30) event in Class I appeared in the cluster 3 (see Figure~\ref{fig:will_distribution}). A nontrivial portion of the Class II events and the Class IV events in the cluster 3, comprising 69\% and 15\%, respectively, of the events in that cluster, had the hysteresis loop pattern of linear-then-clockwise and overlapped between the two classes. In terms of Hamshaw's types, cluster 1 was dominated by Type 2-B (broad clockwise pattern) and cluster 3 was dominated by Type 2-D and Type 4 (narrow linear-then-clockwise pattern). The overlap between Class II and Class IV hysteresis classes is generally accepted in the study of hysteresis loops for watershed, and we also made the same observation from the storm events that belonged to the cluster 3 (see Figure~\ref{fig:av_metadata_cluster_k4}).

\begin{table}[!htb]
\begin{small}
%\ali{reformat based on Scotts suggestion.}
\caption{Distribution by hysteresis loop --- Williams' six classes(upper) and Hamshaw's fifteen types (lower).}\label{tbl:scottDist}
\centering
\resizebox{\textwidth}{!}{%
\begin{tabular}{|p{1.5cm}|c|c|c|c|c|c|c|c|c|c|c|c|c|c|c|c|}
\hline
 & \multicolumn{3}{|c|}{\bf Class I}& \multicolumn{5}{|c|}{\bf Class II} & \multicolumn{3}{|c|}{\bf Class III}& {\bf Class IV} & \multicolumn{2}{|c|}{\bf Class V} & {\bf Complex} & \multirow{2}{*}{\bf Total}\\
\cline{2-15}
\bf{Cluster} & {\bf 1-A}& {\bf 1-B}& {\bf 1-C}& {\bf 2-A}& {\bf 2-B}& {\bf 2-C}& {\bf 2-E}& {\bf 2-D} & {\bf 3-A} &{\bf 3-B} &{\bf 3-C} &{\bf 4} &{\bf 5-A} &{\bf 5-B} & & \\
\thickhline
\multirow{ 2}{*}{1}& \multicolumn{3}{|c|}{11}& \multicolumn{5}{|c|}{167} & \multicolumn{3}{|c|}{16}& \multirow{2}{*}{12} &\multicolumn{2}{|c|}{20}&\multirow{ 2}{*}{8} &\multirow{ 2}{*}{234}\\
\cline{2-12} \cline{14-15}
&6& 4 &  1& 13 & 64& 42& 33  & 15 & 4& 2& 10& & 17& 3 & &\\
\thickhline
\multirow{2}{*}{2}     & \multicolumn{3}{|c|}{12}& \multicolumn{5}{|c|}{58} & \multicolumn{3}{|c|}{16}& \multirow{2}{*}{15} &\multicolumn{2}{c|}{15}&\multirow{ 2}{*}{9} &\multirow{ 2}{*}{125}\\
\cline{2-12} \cline{14-15}
&4& 5 &  3& 12 & 20& 9& 13   &  4   & 4& 10& 2& & 10& 5 & &\\
\thickhline
\multirow{ 2}{*}{3}     & \multicolumn{3}{|c|}{1}& \multicolumn{5}{|c|}{80} & \multicolumn{3}{|c|}{6}& \multirow{2}{*}{18} &\multicolumn{2}{c|}{2}&\multirow{ 2}{*}{9} &\multirow{ 2}{*}{116}\\
\cline{2-12} \cline{14-15}
&0& 1 &  0& 0 & 11& 18& 11   &  40  & 1& 3& 2 & & 2  & 0  & &\\
\thickhline
\multirow{ 2}{*}{4}     & \multicolumn{3}{|c|}{6}& \multicolumn{5}{|c|}{80} & \multicolumn{3}{|c|}{13}& \multirow{2}{*}{14} &\multicolumn{2}{c|}{10}&\multirow{ 2}{*}{5} &\multirow{ 2}{*}{128}\\
\cline{2-12} \cline{14-15}
&2& 3 &  1& 12& 29& 11& 13  & 15 & 2& 5& 6 & & 8  & 2   & &\\
\thickhline
\multirow{2}{*}{\bf Total} &  \multicolumn{3}{|c|}{30}& \multicolumn{5}{|c|}{385} & \multicolumn{3}{|c|}{51}& \multirow{2}{*}{59} &\multicolumn{2}{c|}{47}&\multirow{ 2}{*}{31} &\multirow{ 2}{*}{603}\\
\cline{2-12} \cline{14-15}
 &12 &13 &5 &37 &114 &80 &70 & 74&11 & 20&20 &  &37 &10 & &\\
\thickhline
\end{tabular}
}

\end{small}
\end{table} 

\subsubsection{Relationship to watershed sites} \label{sec:relation_to_watershed}

\begin{table}[!t]
\centering
\caption{Distribution of Mad River watersheds storm events over clusters.}\label{tbl:ClassDist}
\begin{small}
\begin{tabular}{|c|p{1.5cm}|p{1.5cm}|p{2cm}|p{1.5cm}|p{1.5cm}|p{2cm}|c|}
\hline
{\bf Cluster} & {\bf Mad River}& {\bf Shepard Brook}& {\bf High Bridge Brook}& {\bf Mill Brook}& {\bf Folsom Brook}& {\bf Freeman Brook} & {\bf Total}\\
\hline
1     & 44 & 45 & 18 & 68 & 34 & 25 & 234\\
2     & 62 & 18 & 3  & 18 & 12 & 12 & 125 \\
3     & 15 & 23 & 14 & 35 & 20 & 9  & 116\\
4     & 27 & 20 & 6  & 37 & 30 & 8  & 128\\
\hline
{\bf Total} & 148&106 & 41 & 158& 96 & 54 & 603\\
\hline
\end{tabular}
\end{small}

\end{table} 

\begin{figure}[!ht]
\centering
\begin{subfigure}[t]{0.45\textwidth}
\centering
\includegraphics[width=1\textwidth]{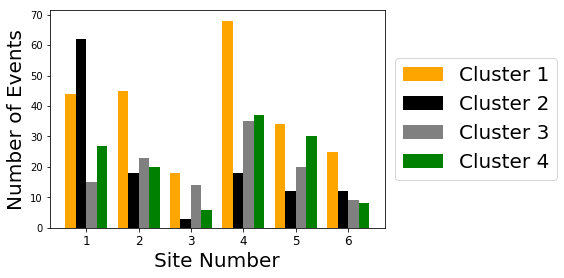}
\caption{Number of storm events at each site.}
\end{subfigure}%
\quad
\begin{subfigure}[t]{0.45\textwidth}
\centering
\includegraphics[width=1\textwidth]{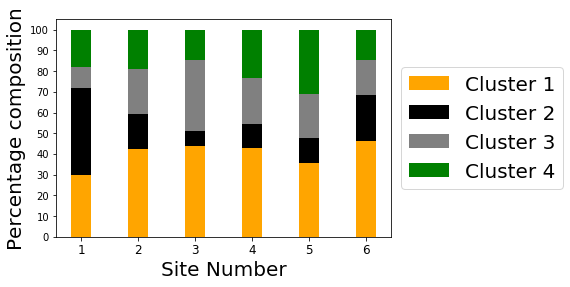}
\caption{Percentage composition of storm events at each site.}
\end{subfigure}\\
\begin{subfigure}[b]{0.45\textwidth}
\centering
\includegraphics[width=1\textwidth]{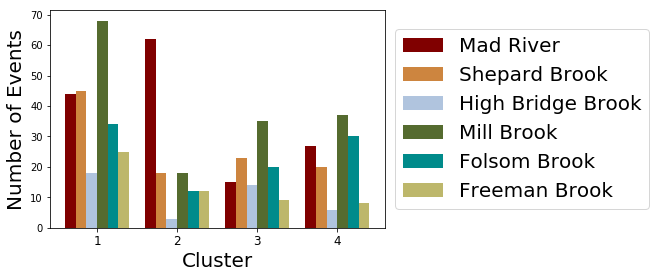}
\caption{Number of storm events in each cluster.}\label{fig:sitec}
\end{subfigure}%
\quad
\begin{subfigure}[b]{0.45\textwidth}
\centering
\includegraphics[width=1\textwidth]{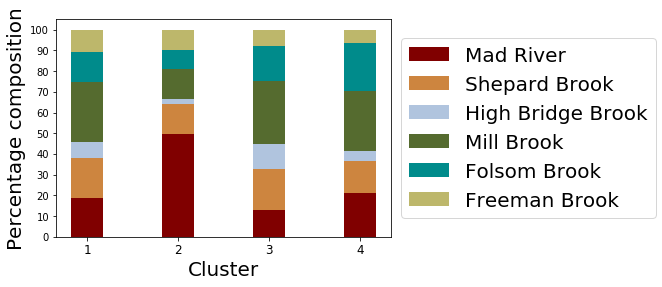}
\caption{Percentage composition of each cluster.}\label{fig:sited}
\end{subfigure}

\caption{Cluster distribution over sites. The distribution of events in clusters is not in-dependant of sites with a p-value of less than 0.001 using a Chi-squared test of independence.}\label{fig:site_distribution}
\end{figure}

Both the physical features of a catchment and characteristics of individual storm events have influence on the type of streamflow and SSC events that occur at a particular monitored site. Figure~\ref{fig:site_distribution} shows the number and percentage of storm events in each cluster from all watersheds in the \cmmnt{primary (603 events)} Mad River dataset. Chi-squared test of independence between watershed sites and clusters had a p-value less than 0.001, thus strongly indicating a correlation between the two. For instance, 42\% of events from the main stem of Mad River watershed site were grouped in the cluster 2 (Figure~\ref{fig:site_distribution}). Moreover, the numbers of storm events from different watersheds in the cluster 2 were in the same order as the stream order of the watersheds and, related, the site 3, the smallest, had the smallest number (only three out of 41) of storm events that appeared in the cluster 2. This observation hints some correlation between the clusters and watershed size.

%\begin{wrapfigure}{l}{0.6\linewidth}
%\end{wrapfigure}

The correlation between watersheds and clusters makes sense given that we could expect a particular site to have a characteristic hydrograph and sedigraph to a certain extent. This can be inferred intuitively when we visually examine the storm evens in the cluster 2, for instance. Cluster 2 contained events with a low degree of discharge recession (i.e., high difference between discharge rates at the start and end, respectively, of the event) as well as low SSC recession. This makes sense given that watersheds with larger catchment areas --- main stem of Mad River watershed has the largest catchment area in our study are --- are likely to take longer to return to the base discharge and concentration quantities and may not completely do so before another event occurs. 

%\scott{Just a side note - the above point highlights our method is sensitive to how well and what method was used to define an event. We'll likely make that point in the discussion as a current limitation and area further research is justified.}\ali{This side note used to be a comment, and now became a part of the paragraph, if intentional let me know <- check with scott)}

\subsubsection{Discrimination of event characteristics through clustering and hysteresis}

Computational clusters differentiated storm events based on a set of storm event metrics that is different from the set of metrics by which storm events in Williams' hysteresis loop classes were differentiated (see Table~\ref{tbl:Anova}). Specifically, for 19 of the 24 storm event metrics, an ANOVA test showed that at least one of the clusters had a mean metric value that was significantly different (i.e., p-value $<$ 0.05) from the mean metric value of rest of the clusters, whereas for 11 metrics, one of the Williams' classes had a mean metric value that was significantly different from the mean metric value of the other Williams' classes. This is not surprising since both methods capture different features of the hydrograph and the sedigraph.

The hysteresis loop classes are designed to extract differences in the timing of the hydrographs and sedigraphs. This difference is the key to reflecting/preserving the shape of the hysteresis loops, and high emphasis is placed on the direction of the loop. Thus, it is not surprising that ANOVA test showed three of the metrics of Table~\ref{tbl:Anova} --- HI (hysteresis index), $T_\textit{PSSC}$ (time between peak SSC and rainfall center of mass), and $T_\textit{QSSC}$ (time between peak SSC and peak flow) --- to have the most explanatory power (as indicated by f-values) for hysteresis loop classes. This would be expected since these three metrics are directly indicative of the timing of SSC in relation to Q, which is key in determining the shape of a hysteresis loop.

In comparison, the explanatory power of clusters for distinguishing characteristics of events was based on a larger and more varied set of metrics (see Table~\ref{tbl:Anova}). In contrast to the hysteresis classification, event clusters had significant differences across metrics associated with hydrograph and sedigraph characteristics --- for instance, the timing of the peak of the hydrograph ($T_Q$) and sedigraph ($T_\textit{SSC}$) as well as the difference between discharge/concentration values at the start of the event and end of the event ($Q_{Recess}$ and $SSC_{Recess}$). This indicates that the clustering of events is driven by the hydrograph and sedigraph themselves as well as the relationship between the two.%\donna{I might be missing something here (and immediately below). Let's talk about it.}\ali{Line immidiately below is commented out now since it wasnt the intend.}
 %
%We additionally note that all performance metrics are significantly explained with the Williams' six hysteresis loop classes are also significantly explained with the four computational clusters, with the exception of $\textit{FLUX}_\textit{NORM}$.\ali{This line makes it a bake-off. We are selecting metrics that are explained by trajectory to begin with. So not a fair comparision. We just want to show that trajectories explain metrics that are not explained by hystersis loop. Hystersis loops explains other metrics that we are not testing because its not a comparision.}
%\byung{Ali, math mode in LaTeX treats $NORM$ in $FLUX_{NORM}$ as multiplication of $N$, $O$, $R$, and $M$ (the same for $FLUX$). We can prevent it by letting LaTeX compiler know that it is a text, for example, like $T_\textit{SSC}$ and $\textit{FLUX}_\textit{NORM}$.} 

%\renewcommand\baselinestretch{1}\selectfont
\begin{table}[!htb]
\centering
\caption{ANOVA test result using watershed performance metrics. F-value is shown in bold when the corresponding p-value is significant (i.e., \cmmnt{significance cutoff p-value} $<$ 0.05) and additionally marked with ``{\bf **}'' if p-value $<$ 0.0001, and with `{\bf *}' if p-value $<$ 0.001.}\label{tbl:Anova}
\newcolumntype{M}[1]{>{\centering\arraybackslash}m{#1}}
\begin{tabular}{|M{2cm}|M{7cm}|M{3.3cm}|M{3.3cm}|}
\hline
\multirow{2}{*}{\bf Metric}&\multirow{2}{*}{\bf Description}&\multicolumn{2}{|c|}{\bf F-value}\\
\cline{3-4}
 &  &  {\bf Hysteresis loop} & {\bf Time series clusters} \\
\hline
\multicolumn{4}{|c|}{\bf Hydrograph/ Sedigraph characteristics}\\
\hline
${T}_{Q}$& Time to peak discharge (hr) & 0.78 &{\bf 50.82**} \\
${T}_{SSC}$& Time to peak TSS (hr) & 1.10 & {\bf 39.69**}\\
${T}_{QSSC}$& Time between peak SSC and peak flow (hr) & {\bf 45.96**} &{\bf 14.02**} \\
$Q_{Recess}$ & Difference in discharge value between the beginning and end of event & {\bf 7.12**} & {\bf 91.64**} \\
$SSC_{Recess}$ & Difference in concentration value between the beginning and end of event & {\bf 19.14**} & {\bf 20.30**}\\
HI& Hysteresis Index &{\bf 283.60**} & {\bf 12.21**}\\
\hline
\multicolumn{4}{|c|}{\bf Antecedent conditions}\\
\hline
$T_{LASTP}$&Time since last event (hr) & 0.46 & 1.92\\
A3P& 3-Day antecedent precipitation (mm) &{\bf 5.61*} & {\bf 12.82**} \\
A14P& 14-Day antecedent precipitation (mm) &{\bf 2.92} & {\bf 7.26*}\\
${SM}_{SHALLOW}$& Antecedent soil moisture at 10 cm depth (\%) & 0.94& 2.08 \\
$SM_{DEEP}$& Antecedent soil moisture at 50 cm depth (\%) &0.74 &  1.19\\
$BF_{NORM}$& Drainage area normalized pre-storm baseflow ($m^3/s/km^2)$ &0.36 & {\bf 5.46}\\
\hline
\multicolumn{4}{|c|}{\bf Rainfall characteristics}\\
\hline
P& Total event precipitation (mm) &{\bf 4.01} & {\bf 10.72**}\\
$P_{max}$ & Maximum rainfall intensity (mm)  & 2.48 & {\bf 20.81**}\\
$D_{P}$& Duration of precipitation (hr) & 1.68 &{\bf 11.36**} \\
${T}_{PSSC}$& Time between peak SSC and rainfall center of mass (hr)  &{\bf 60.34**} & {\bf 22.69**}\\
\hline
\multicolumn{4}{|c|}{\bf Streamflow and sediment characteristics}\\
\hline
BL & Basin Lag & {\bf 6.03*} & \bf{21.08**}\\
${Q}_{NORM}$& Drainage area normalized stormflow ($m^3 / s / km^2$) & 1.30& {\bf 5.60*}\\
Log(${Q}_{NORM}$)& Log-normal stormflow quantile (\%) & {\bf 4.95} & {\bf 25.10**}\\
${D}_{Q}$& Duration of stormflow (hr) & 0.43 & {\bf 9.50**}\\
FI& Flood intensity &2.25  & {\bf 8.90**}\\
SSC& Peak SSC (mg/L) & 0.74 & {\bf 8.63**} \\
${SSL}_{NORM}$& Drainage area normalized total sediment ($kg/m^2$) & 2.50 & 1.76 \\
${FLUX}_{NORM}$& Drainage area and flow normalized sediment flux ($kg/m^3/km^2$) &{\bf 10.52**} & 0.18\\
\hline

\end{tabular}
\end{table}

\subsubsection{Characteristics of event clusters}

\begin{figure}[H]
\centering
\hspace*{-0.225in}
\includegraphics[width=1\textwidth]{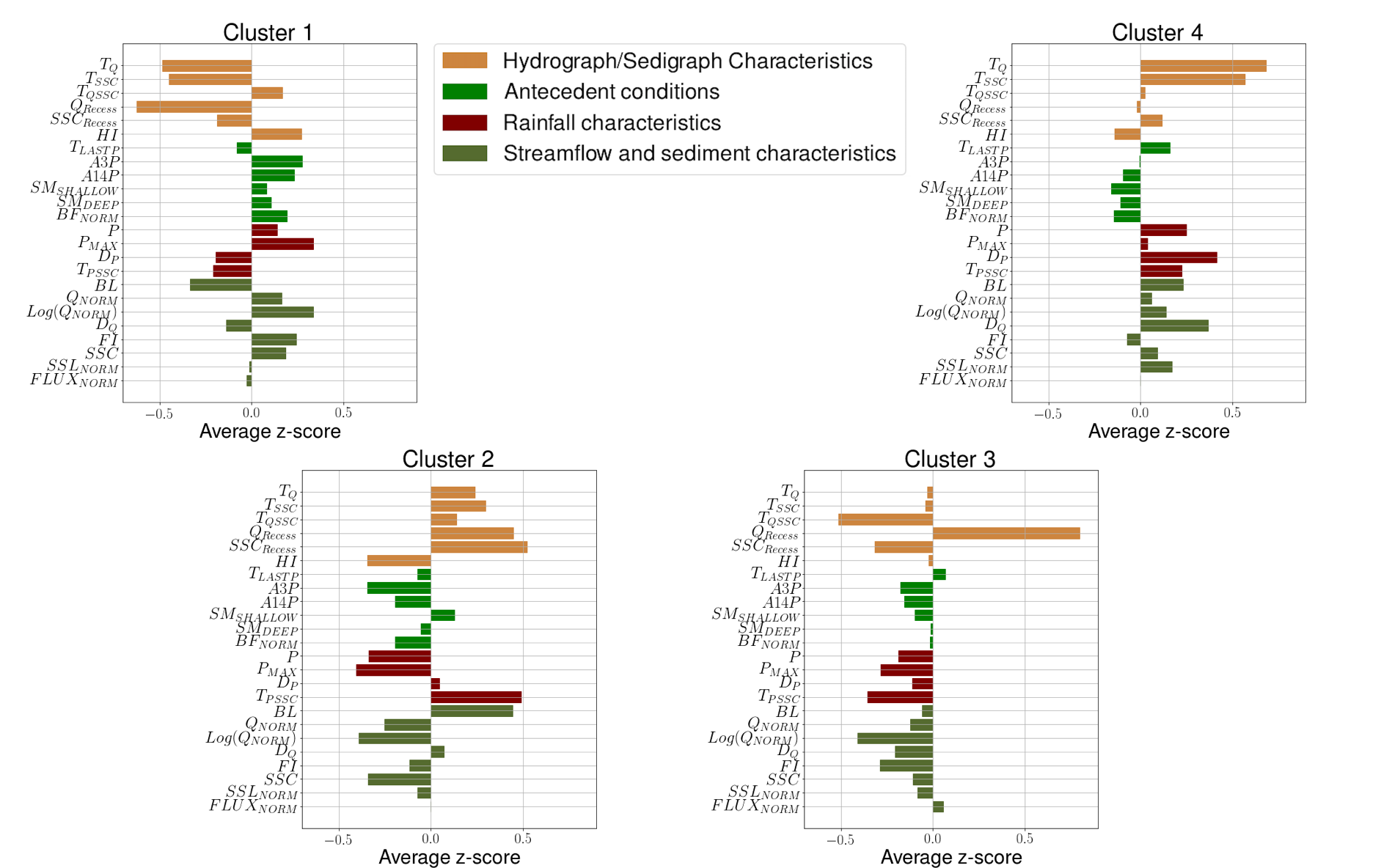}
\caption{Average $z$-scores of storm event metrics for each of the four clusters.%\byung{Let's show these subfigures in two rows, two in each row; the legend box can be thrown into the bottom or top of the figure, I think.}
}\label{fig:zscores}
\end{figure}

Each of the four clusters exhibited certain characteristics of events, observed from the event visualizations (see Figure~\ref{fig:av_metadata_cluster_k4}) and the z-score value of storm event metrics (see Figure~\ref{fig:zscores}). The metrics were used not as input to the clustering algorithm but as means to study the characteristics of resulting clusters. %and validate the efficacy of the suggested clustering method.

\begin{comment}
\begin{verbatim}
Characteristics of cluster 1 events:
- wet antecedent conditions (high $BF_{Norm}$,
- slow recession of streamflow (Figure 10a)
- early sediment pulse (Figure 10a)
- broad clockwise hysteresis pattern (Figure 10a and high HI)
 $SM_{Deep}$, $SM_{Shallow}$, $A3P$ and $A14P$)
- high stream flow (high $Log(Q_{Norm})$, $Q_{Norm}$, and $FI$)
- large precipitation (high $P$ and $P_{Max}$)
- high SSC (high SSC)
- quick rising SSC (low $T_{SSC}$)
- quick rising streamflow (low $T_{Q}$)
- streamflow returns to baseflow (low $Q_{Recess}$)
- sedimentflow returns to baseflow (low $SSC_{Recess}$)
\end{verbatim}
\end{comment}

Storm events in the cluster 1 were larger events with wetter antecedent conditions that result in higher streamflows and higher SSC. These characteristics are based on the following observations. First, the events have larger amount of precipitation (positive z-score for $P$ and $P_\textit{Max}$) resulting in larger streamflows (positive z-score for Log $(Q_\textit{Norm})$, $Q_\textit{Norm}$, and $FI$). Second, the events have positive z-score for $BF_\textit{Norm}$, $SM_\textit{Deep}$,$SM_\textit{Shallow}$, $A3P$ and $A14P$, which characterize wetter antecedent conditions.
Other key characteristics are that their hydrographs do not return quickly to baseline flow, sedigraph pulses occur early, and the dominating hysteresis shape is a broad clockwise pattern (see Figure~\ref{fig:av_metadata_cluster_k4_1}). %\donna{Again, I'd love to see the breakdown before we talk too much about hysteresis direction.} 
The hysteresis pattern is also confirmed from positive z-score of HI. %\donna{This if the metric that confuses me a bit. I think we have to figure out how to describe what the HI actually captures (compared with the hysteresis loop itself).}
Additional event characteristics include negative z-score of $T_\textit{SSC}$ and $T_\textit{Q}$, meaning quickly rising sedigraph and hydrograph, respectively, and negative z-score of $Q_\textit{Recess}$ and $SSC_\textit{Recess}$, meaning that streamflow and SSC, respectively, return to base levels at the end of the events.

\begin{comment}
\begin{verbatim}
- lesser precipitation (PMax, P)
- smaller streamflow (Log(Q_Norm), QNorm, FI)
- streamflow does not return to baseflow (High Q Recess)
- SSC does not return to baseflow (High SSC Recess)
- long time from peak discharge to peak SSC (T_PSSC)
- Lower HI value - HI index
- drier antecedent conditions (BF_Norm, A3P, A14P)
- Lower peak SSC (SSC)

- Long BL ?
\end{verbatim}
\end{comment}

Cluster 2 included smaller storm events with drier antecedent conditions in which the streamflow and SSC do not return to base levels at the end of the events.  These characteristics are based on the following observations.
First, the events have smaller amount of precipitation (negative z-score of $P$ and $P_\textit{Max}$), resulting in smaller streamflows (negative z-score of Log $(Q_\textit{Norm})$, $Q_\textit{Norm}$, and $FI$). Second, the events have negative z-score values for $BF_\textit{Norm}$, $SM_\textit{Deep}$, $SM_\textit{Shallow}$, $A3P$ and $A14P$, which characterize drier antecedent conditions. Third, the events have positive z-score values for $Q_\textit{Recess}$ and $SSC_\textit{Recess}$, meaning that the streamflow and the SSC do not return to base levels at the end of the events (see Figure~\ref{fig:av_metadata_cluster_k4_2}). 
Other key characteristics include a longer time from peak SSC to peak rainfall center of mass (positive z-score of $T_\textit{PSSC}$) and that the dominating hysteresis shape is a narrow loop (see Figure~\ref{fig:av_metadata_cluster_k4_2}). The hysteresis pattern is also confirmed from negative z-score of hysteresis index.
Additional characteristics include a negative z-score of $SSC$, meaning a lower peak SSC amount, and a negative z-score of $BL$, meaning that the watersheds respond slowly to a rainfall event.

\begin{comment}
\begin{verbatim}
- smaller precipitation (P, Pmax)
- smaller streamflows (Log(Q_Norm), QNorm, FI)
- Average antecident conditions (zero $BF_{Norm}$, $SM_{Deep}$,$SM_{Shallow}$, $A3P$ and $A14P$)
- streamflow does not return back to baseflow (high Q recess)
- SSC lingers or is multi-peaked ( SSC Recess, figure)

- short time to peak between SSC and peak flow (T {QSSC})
- short time from peak discharge to peak SSC (small T_PSSC)
- Streamflow has a fast rate of rise (figure)
- SSC has a fast rate of rise (figure)
\end{verbatim}
\end{comment}

Cluster 3 included smaller events that occurred in average antecedent conditions in which streamflow does not completely return to baseflow and SSC also lingers above base-level or is multi-peaked. These characteristics are based on the following observations.
First, the events have smaller amount of precipitation (negative z-score of $P$ and $P_\textit{Max}$), resulting in smaller streamflows (negative z-score of Log $(Q_\textit{Norm})$, $Q_\textit{Norm}$, and $FI$). Second, the events have near zero z-score values for $BF_\textit{Norm}$, $SM_\textit{Deep}$,$SM_\textit{Shallow}$, $A3P$ and $A14P$, which characterize average antecedent conditions. Third, the events have a positive z-score for $Q_\textit{Recess}$ and negative z-score for $SSC_\textit{Recess}$, meaning that the streamflow does not completely return to baseflow while SSC does. It can be observed from Figure~\ref{fig:av_metadata_cluster_k4_3} that SSC lingers or is multi-peaked.
Other key characteristics include a short time from peak SSC to peak rainfall center of mass (negative z-score of $T_\textit{PSSC}$) and short time from peak discharge to peak SSC (negative z-score of $T_\textit{QSSC}$).
Additional characteristics include a fast rate of rise for both streamflow and SSC (see Figure~\ref{fig:av_metadata_cluster_k4_3}).

\begin{comment}
\begin{verbatim}
- longer events (D_Q)
- Long duration P (D_P)
- High total rainfall (Higher P)
- Low maximum rainfall quantity (low P_max)
- average to slightly dry antecedent conditions ($BF_{Norm}$, $SM_{Deep}$,$SM_{Shallow}$, $A3P$ and $A14P$)
- longer time to peak for SSC (T_SSC)
- longer time to peak for Q (T_Q)

- Higher SSL loading (Higher SSL_NORM)?
- Long basin Lag (BL)?
\end{verbatim}
\end{comment}

Cluster 4 included longer and less intense events occurring during average to slightly dry antecedent conditions. These characteristics are based on the following observations.
First, the events have longer duration (positive z-score of $D_Q$) and high total precipitation (positive z-score of $P$) but low maximum precipitation (negative z-score of $P_\textit{Max}$) resulting in near average streamflows (near zero z-score for Log $(Q_\textit{Norm})$, $Q_\textit{Norm}$, and $FI$). Second, the events have slightly negative z-score values for $BF_\textit{Norm}$, $SM_\textit{Deep}$, $SM_\textit{Shallow}$, $A3P$ and $A14P$, which characterize average to slightly dry antecedent conditions. 
Other key characteristics include long time to peak SSC and Q (positive z-score for $T_\textit{SSC}$ and $T_Q$).
Additional characteristics include a larger amount of sediments transported during an event (positive $SSL_\textit{Norm}$).

\section{Discussion}

\subsection{Hydrological implications of the results}

%\donna{Given that many of our "results" figures are in the processes of changing, I'm going to hold off on editing the discussion. Scott and I have talked a little about this; but it would be good for us to all be in the same room in the next week or two. Ali turns around the new reuslts/figures fast enough that I think the story should converge pretty quickly. I like where this is going...very fun.}  
%\scott{think about referencing specifically a couple of sample events where hydrograph/sedigraph looks similar but hysteresis is very different}
%The main conclusion of the study is that the computational clustering is a comprehensive approach to categorizing storm events in the following regards --- (i) computational clusters show significant level of relationships to hysteresis loop classes as well as to watershed sites while not being similar and (ii) computational clusters explain a different set of storm event metrics in comparison to hysteresis loop classes.
Main implications we can draw from the results are that in the Mad River watershed (i) the optimal number of categories of storm events is four (see Figure~\ref{fig:best_k_av}); (ii) events in computational clusters have significant relationships to events in hysteresis loop classes and events in watershed sites, while not identical (see Figure~\ref{fig:will_distribution} and Figure~\ref{fig:site_distribution}); (iii) metrics that differentiate events in computational clusters are different from metrics that differentiate events in hysteresis loop classes (see Table~\ref{tbl:Anova}); (iv) events in different computational clusters are significantly different in terms of metrics associated with hydrograph and sedigraph characteristics (see Table~\ref{tbl:Anova}); and (v) events in each cluster share certain unique characteristics in terms of all 24 storm event metrics (see Figure~\ref{fig:zscores}).

The results also suggest that the computational clustering approach identified events caused by sediment delivery from upstream sources. Hysteresis approach is typically used only for small sized watersheds (smaller than 100 $km^2$) since large watershed are affected by sediment delivery from upstream sources~\citep{Hamshaw2018}.  In our results, however, the cluster 2 is dominated by events from the main stem of Mad River watershed, while events in the cluster 2 have smaller precipitations and stream flow and sediment flow do not come back to the base level. These are indicative of consistent sediment and streamflow delivery from upstream sources that  might be experiencing or might have experienced a storm event shortly before. Additionally, streamflow not returning to the base level at the end of an event is likely to indicate saturated soils, high groundwater tables or soil moisture, a feature that the current classification schemes based on hysteresis shape/direction cannot capture. Since the computational clustering captured this difference in discharge values at the start and end of the event, it could be a promising addition to methods used to create an early warning system for impending floods.

%\subsection{Broader applications of the proposed computational method}

%\byung{Scott, will you be able to fill in this subsection? I would love to brainstorm it with you if you would. Please feel free to change the organization of this section as you see fit.}\scott{yes, I can. Will want to get across the applicability to 1, categorization of events, 2, hydrological model calibration, and 3, comparison of events/watersheds for similarity}

\subsection{Challenges and opportunities}

%The storm event analysis presented in this work is a novel application for multivariate time series (MTS) clustering. %, which is an advancement over the current hysteresis loop approach. 
%It brings some new challenges and opportunities.
%\scott{Remove sentence above and just start with 5.3 - Challenges}
%\scott{minor item - but maybe start with challenges, and then have opportunities?}\ali{done}

%\subsection{Challenges}

The sparsity of data is an inherent challenges in storm event analysis. Our study area --- a typical humid and temperate watershed --- experiences only about 30 storm events a year. Other recent prominent event-based studies~\citep{Wymore2019,Sherriff_etal_2016,Vaughan_etal_WRR_2017} featured between 8 and 90 events per site monitored. This inherent sparsity of data is compounded when analyzing multivariate time series generated from sensors, as all sensors of different modality must be online and operational simultaneously, which is a significant challenge in \emph{in-situ} water quality monitoring. Besides, increased dimensionality (i.e., number of variables) of data would cause storm events to be even sparser % along the added dimensions. This challenge, often faced when increasing the dimensionality of data space, is 
(called the ``curse of dimensionality''~\citep{Bellman1957}). %--- that is, when the dimensionality of data space increases, the volume of the data space increases much more, so that data in the enlarged space becomes sparser. 
 %
%In addition, the small number of events considered in typical event-based studies in hydrology is partially responsible for the reliance on manual classification and for lack of advancement in identifying new clusters or categories of events. To train more advanced computation models, such as deep neural networks and to identify new categories of event responses requires compiling larger datasets, and hence the advancement in these areas is constrained because of data sparsity~\citep{Hamshaw2018}. Efforts are being made in the field of hydrology to compile larger datasets across researchers and organizations~\citep{CUAHSI} to facilitate training of more advanced computation models and identify new categories of event responses. The approach of generating synthetic storm event discharge and concentration time series introduced in this work is another method that can used for training models in order to overcome data sparsity issues. Overall, with the advent of the big data era, the advancement in sensor technology, availability of synthetic dataset generation methodology, and community efforts to compile larger datasets, now is the opportune time for us to step into the higher dimensional analysis for storm events.
Currently, efforts are being made in the field of hydrology to compile larger datasets across researchers and organizations~\citep{CUAHSI} to address the data sparsity issue. Generating synthetic storm events as was done in our work could be another approach.

Determining the optimal value of $k$, the number of clusters of storm events, is challenging in a hydrological application like ours where there is not always clear separation of groups. Using the elbow method can be subjective, and sometimes the characteristic elbow is not clearly visible in an elbow plot.   %The synthetic data generated in this study was well separated and the elbow plot for synthetic data showed a clearer elbow in comparison to the elbow plots for real datasets used in this study.
If identifying the elbow becomes problematic, we may consider different options for the analysis steps such as preprocessing, distance measure, and clustering algorithm. Regarding the clustering algorithm for example, a density-based clustering algorithm~\citep{Ester1996}, which does not require the number of clusters as an input, can be considered.  

The computational clustering approach used in this work is also applicable to other solutes (or constituents) that demonstrate patterns different from those that are observed in SSC~\citep{Lloyd2016,Zuecco2016}. 
%\byung{Elaborate.  Give concrete examples of other solutes, and examples of probably patterns for them.}
 %
Moreover, the computational approach can be extended beyond using a single solute (e.g., SSC) to using multiple solutes (e.g., SSC, phosphorous, CO2) together in order to reveal any unknown interactions among them in watershed events.

\bibliographystyle{apalike}
\bibliography{main.bib}

\begin{thebibliography}{}

\bibitem[Aguilera and Melack, 2018]{Aguilera_Melack_2018}
Aguilera, R. and Melack, J.~M. (2018).
\newblock Concentration\nobreakhyphen discharge responses to storm events in
  coastal california watersheds.
\newblock {\em Water Resources Research}, 54(1):407–424.

\bibitem[{Banerjee} and {Dave}, 2004]{Banerjee_2004}
{Banerjee}, A. and {Dave}, R.~N. (2004).
\newblock Validating clusters using the hopkins statistic.
\newblock In {\em Proceedings of the 2004 IEEE International Conference on
  Fuzzy Systems (IEEE Cat. No.04CH37542)}, volume~1, pages 149--153 vol.1.

\bibitem[Begum et~al., 2015]{Nurjahan2016}
Begum, N., Ulanova, L., Wang, J., and Keogh, E. (2015).
\newblock Accelerating dynamic time warping clustering with a novel admissible
  pruning strategy.
\newblock In {\em Proceedings of the 21th ACM SIGKDD International Conference
  on Knowledge Discovery and Data Mining}, KDD '15, pages 49--58, New York, NY,
  USA. ACM.

\bibitem[Bellman, 1957]{Bellman1957}
Bellman, R. (1957).
\newblock {\em {Dynamic Programming}}.
\newblock Dover Publications.

\bibitem[Bende-Michl et~al., 2013]{Bende-Michl2013}
Bende-Michl, U., Verburg, K., and Cresswell, H.~P. (2013).
\newblock High-frequency nutrient monitoring to infer seasonal patterns in
  catchment source availability, mobilisation and delivery.
\newblock {\em Environmental Monitoring and Assessment}, 185(11):9191--9219.

\bibitem[Burns et~al., 2019]{Burns_etal_2019}
Burns, D.~A., Pellerin, B.~A., Miller, M.~P., Capel, P.~D., Tesoriero, A.~J.,
  and Duncan, J.~M. (2019).
\newblock Monitoring the riverine pulse: Applying high-frequency nitrate data
  to advance integrative understanding of biogeochemical and hydrological
  processes.
\newblock {\em Wiley Interdisciplinary Reviews: Water}, page e1348.

\bibitem[Burt et~al., 2015]{Burt_etal_HP_2015}
Burt, T.~P., Worrall, F., Howden, N. J.~K., and Anderson, M.~G. (2015).
\newblock Shifts in discharge-concentration relationships as a small catchment
  recover from severe drought.
\newblock {\em Hydrological Processes}, 29(4):498–507.

\bibitem[Chen et~al., 2017]{Chen_etal_JoH_2017}
Chen, L., Sun, C., Wang, G., Xie, H., and Shen, Z. (2017).
\newblock Event-based nonpoint source pollution prediction in a scarce data
  catchment.
\newblock {\em Journal of Hydrology}, 552(Supplement C):13–27.

\bibitem[CUAHSI, 2019]{CUAHSI}
CUAHSI (2019).
\newblock Consortium of universities for the advancement of hydrologic science,
  inc.
\newblock \url{https://www.cuahsi.org}.

\bibitem[Dau et~al., 2018]{UCRArchive2018}
Dau, H.~A., Keogh, E., Kamgar, K., Yeh, C.-C.~M., Zhu, Y., Gharghabi, S.,
  Ratanamahatana, C.~A., Yanping, Hu, B., Begum, N., Bagnall, A., Mueen, A.,
  and Batista, G. (2018).
\newblock The {UCR} time series classification archive.
\newblock \url{https://www.cs.ucr.edu/~eamonn/time_series_data_2018/}.

\bibitem[Ehret and Zehe, 2011]{hess2011}
Ehret, U. and Zehe, E. (2011).
\newblock Series distance - an intuitive metric to quantify hydrograph
  similarity in terms of occurrence, amplitude and timing of hydrological
  events.
\newblock {\em Hydrology and Earth System Sciences}, 15(3):877--896.

\bibitem[Ester et~al., 1996]{Ester1996}
Ester, M., Kriegel, H.-P., Sander, J., and Xu, X. (1996).
\newblock A density-based algorithm for discovering clusters a density-based
  algorithm for discovering clusters in large spatial databases with noise.
\newblock In {\em Proceedings of the Second International Conference on
  Knowledge Discovery and Data Mining}, KDD'96, pages 226--231. AAAI Press.

\bibitem[Ewen, 2011]{EWEN2011178}
Ewen, J. (2011).
\newblock Hydrograph matching method for measuring model performance.
\newblock {\em Journal of Hydrology}, 408(1):178 -- 187.

\bibitem[Hamshaw et~al., 2018]{Hamshaw2018}
Hamshaw, S., M.~Dewoolkar, M., W.~Schroth, A., Wemple, B., and M.~Rizzo, D.
  (2018).
\newblock A new machine-learning approach for classifying hysteresis in
  suspended-sediment discharge relationships using high-frequency monitoring
  data.
\newblock {\em Water Resources Research}, 54.

\bibitem[Javed, 2019a]{dtw_d_code}
Javed, A. (2019a).
\newblock Dynamic time warping.
\newblock \url{https://github.com/ali-javed/dynamic-time-warping}.

\bibitem[Javed, 2019b]{multi_kmed}
Javed, A. (2019b).
\newblock Multivariate time series dynamic time warping using euclidean
  distance.
\newblock \url{https://github.com/ali-javed/Multivariate-Kmedoids}.

\bibitem[Jones et~al., 2011]{Jones2011}
Jones, A.~S., Stevens, D.~K., Horsburgh, J.~S., and Mesner, N.~O. (2011).
\newblock Surrogate measures for providing high frequency estimates of total
  suspended solids and total phosphorus concentrations1.
\newblock {\em JAWRA Journal of the American Water Resources Association},
  47(2):239--253.

\bibitem[Keesstra et~al., 2019]{Keesstra_etal_2019}
Keesstra, S.~D., Davis, J., Masselink, R.~H., Casalí, J., Peeters, E. T.
  H.~M., and Dijksma, R. (2019).
\newblock Coupling hysteresis analysis with sediment and hydrological
  connectivity in three agricultural catchments in navarre, spain.
\newblock {\em Journal of Soils and Sediments}, 19(3):1598–1612.

\bibitem[{Latecki} et~al., 2005]{Latecki2005}
{Latecki}, L.~J., {Megalooikonomou}, V., {Qiang Wang}, {Lakaemper}, R.,
  {Ratanamahatana}, C.~A., and {Keogh}, E. (2005).
\newblock Partial elastic matching of time series.
\newblock In {\em Proceedings of the 5th IEEE International Conference on Data
  Mining (ICDM'05)}, pages 4 pp.--.

\bibitem[Latecki et~al., 2005]{Latecki2005-2}
Latecki, L.~J., Megalooikonomou, V., Wang, Q., Lakaemper, R., Ratanamahatana,
  C.~A., and Keogh, E. (2005).
\newblock Elastic partial matching of time series.
\newblock In Jorge, A.~M., Torgo, L., Brazdil, P., Camacho, R., and Gama, J.,
  editors, {\em Knowledge Discovery in Databases: PKDD 2005}, pages 577--584,
  Berlin, Heidelberg. Springer Berlin Heidelberg.

\bibitem[Lloyd et~al., 2016a]{Lloyd2016}
Lloyd, C., Freer, J., Johnes, P., and Collins, A. (2016a).
\newblock Using hysteresis analysis of high-resolution water quality monitoring
  data, including uncertainty, to infer controls on nutrient and sediment
  transfer in catchments.
\newblock {\em Science of The Total Environment}, 543, Part A:388 -- 404.

\bibitem[Lloyd et~al., 2016b]{Lloyd2016a}
Lloyd, C. E.~M., Freer, J.~E., Johnes, P.~J., and Collins, A.~L. (2016b).
\newblock Technical note: Testing an improved index for analysing storm
  discharge--concentration hysteresis.
\newblock {\em Hydrology and Earth System Sciences}, 20(2):625--632.

\bibitem[Mather and Johnson, 2015]{MATHER2015}
Mather, A.~L. and Johnson, R.~L. (2015).
\newblock Event-based prediction of stream turbidity using a combined cluster
  analysis and classification tree approach.
\newblock {\em Journal of Hydrology}, 530:751 -- 761.

\bibitem[Minaudo et~al., 2017]{Minaudo2017}
Minaudo, C., Dupas, R., Gascuel-Odoux, C., Fovet, O., Mellander, P.-E., Jordan,
  P., Shore, M., and Moatar, F. (2017).
\newblock Nonlinear empirical modeling to estimate phosphorus exports using
  continuous records of turbidity and discharge.
\newblock {\em Water Resources Research}, 53.

\bibitem[Onderka et~al., 2012]{Onderka2012}
Onderka, M., Krein, A., Wrede, S., Martinez-Carreras, N., and Hoffmann, L.
  (2012).
\newblock Dynamics of storm-driven suspended sediments in a headwater catchment
  described by multivariable modeling.
\newblock {\em Journal of Soils and Sediments}, 12(4):620--635.

\bibitem[Paparrizos and Gravano, 2016]{Paparrizos2016}
Paparrizos, J. and Gravano, L. (2016).
\newblock K-shape: Efficient and accurate clustering of time series.
\newblock {\em SIGMOD Record}, 45(1):69--76.

\bibitem[Paparrizos and Gravano, 2017]{Paparrizos2017}
Paparrizos, J. and Gravano, L. (2017).
\newblock Fast and accurate time-series clustering.
\newblock {\em ACM Transactions on Database Systems}, 42(2):8:1--8:49.

\bibitem[PRISM, 2019]{PRISM2015}
PRISM (2019).
\newblock {PRISM} climate group.
\newblock \url{http://prism.oregonstate.edu}.
\newblock Last accessed on March 16, 2019.

\bibitem[Rakthanmanon et~al., 2012]{Rakthanmanon2012}
Rakthanmanon, T., Campana, B., Mueen, A., Batista, G., Westover, B., Zhu, Q.,
  Zakaria, J., and Keogh, E. (2012).
\newblock Searching and mining trillions of time series subsequences under
  dynamic time warping.
\newblock In {\em Proceedings of the 18th ACM SIGKDD International Conference
  on Knowledge Discovery and Data Mining}, pages 262--270.

\bibitem[Ratanamahatana and Keogh, 2004]{chotirat2004}
Ratanamahatana, C.~A. and Keogh, E. (2004).
\newblock Everything you know about dynamic time warping is wrong.
\newblock In {\em Proceedings of the 3rd Workshop on Mining Temporal and
  Sequential Data}. Citeseer.

\bibitem[Rose et~al., 2018]{Rose_etal_HP_2018}
Rose, L.~A., Karwan, D.~L., and Godsey, S.~E. (2018).
\newblock Concentration\nobreakhyphen discharge relationships describe solute
  and sediment mobilization, reaction, and transport at event and longer
  timescales.
\newblock {\em Hydrological Processes}, 32(18):2829–2844.

\bibitem[Rosenberg and Hirschberg, 2007]{Rosenberg2017}
Rosenberg, A. and Hirschberg, J. (2007).
\newblock V-measure: A conditional entropy-based external cluster evaluation
  measure.
\newblock In {\em Proceedings of the 2007 Joint Conference on Empirical Methods
  in Natural Language Processing and Computational Natural Language Learning},
  pages 410--420.

\bibitem[{Sakoe} and {Chiba}, 1978]{Sakoe1978}
{Sakoe}, H. and {Chiba}, S. (1978).
\newblock Dynamic programming algorithm optimization for spoken word
  recognition.
\newblock {\em IEEE Transactions on Acoustics, Speech, and Signal Processing},
  26(1):43--49.

\bibitem[{Satopaa} et~al., 2011]{Satopaa2011}
{Satopaa}, V., {Albrecht}, J., {Irwin}, D., and {Raghavan}, B. (2011).
\newblock Finding a "kneedle" in a haystack: Detecting knee points in system
  behavior.
\newblock In {\em Proceedings of the 31st International Conference on
  Distributed Computing Systems Workshops}, pages 166--171.

\bibitem[Scipy, 2012]{Savitzky-Golay}
Scipy (2012).
\newblock Savitzky golay filtering.
\newblock \url{https://scipy-cookbook.readthedocs.io/items/SavitzkyGolay.html}.
\newblock Last accessed on February 13, 2019.

\bibitem[Scipy, 2019]{Univariate-Spline}
Scipy (2019).
\newblock Savitzky golay filtering.
\newblock
  \url{https://docs.scipy.org/doc/scipy/reference/generated/scipy.interpolate.UnivariateSpline.html}.
\newblock Last accessed on February 13, 2019.

\bibitem[Sherriff et~al., 2016]{Sherriff_etal_2016}
Sherriff, S.~C., Rowan, J.~S., Fenton, O., Jordan, P., Melland, A.~R.,
  Mellander, P.-E., and hUallachain, D.~O. (2016).
\newblock Storm event suspended sediment-discharge hysteresis and controls in
  agricultural watersheds: Implications for watershed scale sediment
  management.
\newblock {\em Environmental Science \& Technology}, 50(4):1769–1778.

\bibitem[Shokoohi-Yekta and Keogh, 2015]{ShokoohiYekta2015}
Shokoohi-Yekta, M. and Keogh, E.~J. (2015).
\newblock On the non-trivial generalization of dynamic time warping to the
  multi-dimensional case.
\newblock In {\em Proceedings of the 2015 SIAM International Conference on Data
  Mining}.

\bibitem[Stryker et~al., 2017]{Stryker2017}
Stryker, J., Wemple, B., and Bomblies, A. (2017).
\newblock Modeling sediment mobilization using a distributed hydrological model
  coupled with a bank stability model.
\newblock {\em Water Resources Research}, 53(3):2051--2073.

\bibitem[Vaughan et~al., 2017]{Vaughan_etal_WRR_2017}
Vaughan, M. C.~H., Bowden, W.~B., Shanley, J.~B., Vermilyea, A., Sleeper, R.,
  Gold, A.~J., Pradhanang, S.~M., Inamdar, S.~P., Levia, D.~F., Andres, A.~S.,
  and et~al. (2017).
\newblock High-frequency dissolved organic carbon and nitrate measurements
  reveal differences in storm hysteresis and loading in relation to land cover
  and seasonality: high-resolution doc and nitrate dynamics.
\newblock {\em Water Resources Research}, 53(7).

\bibitem[Wemple et~al., 2017]{Wemple2017}
Wemple, B.~C., Clark, G.~E., Ross, D.~S., and Rizzo, D.~M. (2017).
\newblock Identifying the spatial pattern and importance of hydro-geomorphic
  drainage impairments on unpaved roads in the northeastern usa.
\newblock {\em Earth Surface Processes and Landforms}, 42(11):1652--1665.

\bibitem[Wendi et~al., 2019]{Wendi2019}
Wendi, D., Merz, B., and Marwan, N. (2019).
\newblock Assessing hydrograph similarity and rare runoff dynamics by cross
  recurrence plots.
\newblock {\em Water Resources Research}, 55(6):4704--4726.

\bibitem[Wikipedia, 2019]{syntheticData}
Wikipedia (2019).
\newblock Synthetic data.
\newblock \url{https://en.wikipedia.org/wiki/Synthetic_data}.

\bibitem[Williams, 1989]{WILLIAMS1989}
Williams, G.~P. (1989).
\newblock Sediment concentration versus water discharge during single
  hydrologic events in rivers.
\newblock {\em Journal of Hydrology}, 111(1):89 -- 106.

\bibitem[Williams et~al., 2018]{Williams_etal_JoH_2018}
Williams, M.~R., Livingston, S.~J., Penn, C.~J., Smith, D.~R., King, K.~W., and
  Huang, C.-h. (2018).
\newblock Controls of event-based nutrient transport within nested headwater
  agricultural watersheds of the western lake erie basin.
\newblock {\em Journal of Hydrology}, 559:749–761.

\bibitem[Wu et~al., 2007]{Wu2007}
Wu, X., Kumar, V., Ross~Quinlan, J., Ghosh, J., Yang, Q., Motoda, H.,
  McLachlan, G.~J., Ng, A., Liu, B., Yu, P.~S., Zhou, Z.-H., Steinbach, M.,
  Hand, D.~J., and Steinberg, D. (2007).
\newblock Top 10 algorithms in data mining.
\newblock {\em Knowledge and Information Systems}, 14(1):1--37.

\bibitem[Wymore et~al., 2019]{Wymore2019}
Wymore, A.~S., Leon, M.~C., Shanley, J.~B., and McDowell, W.~H. (2019).
\newblock Hysteretic response of solutes and turbidity at the event scale
  across forested tropical montane watersheds.
\newblock {\em Frontiers in Earth Science}, 7:126.

\bibitem[Zuecco et~al., 2016]{Zuecco2016}
Zuecco, G., Penna, D., Borga, M., and van Meerveld, H.~J. (2016).
\newblock A versatile index to characterize hysteresis between hydrological
  variables at the runoff event timescale.
\newblock {\em Hydrological Processes}, 30(9):1449--1466.

\end{thebibliography}

\end{document}